\documentclass[journal]{IEEEtran}

\usepackage{cite}
\usepackage{amsmath,amssymb,amsfonts}
\usepackage{algorithmic}
\usepackage{graphicx}
\usepackage{textcomp}
\usepackage{booktabs}
\usepackage{xcolor}
\usepackage{url}
\usepackage{tikz}
\usetikzlibrary{positioning,arrows.meta}

\usepackage{bm}

\newcommand{\vect}[1]{\mathbf{#1}}
\newcommand{\mat}[1]{\mathbf{#1}}

\def\BibTeX{{\rm B\kern-.05em{\sc i\kern-.025em b}\kern-.08em
    T\kern-.1667em\lower.7ex\hbox{E}\kern-.125emX}}

\begin{document}

\title{Vision-Based Hand Shadowing for Robotic Manipulation via Inverse Kinematics}

\author{Hendrik~Chiche,
        Antoine~Jamme,
        Trevor~Rigoberto~Martinez,
        and~Gabriel~Gomes%
\thanks{H.~Chiche and A.~Jamme are with OMGrab Inc.\ and affiliated with the
University of California, Berkeley, CA 94720 USA, through the Capstone Project
program of the Fung Institute for Engineering Leadership
(e-mail: hendrik\_chiche@berkeley.edu; antoine\_jamme@berkeley.edu).}%
\thanks{T.~R.~Martinez and G.~Gomes are with the Department of Mechanical
Engineering, University of California, Berkeley, CA 94720 USA
(e-mail: tre\_mart@berkeley.edu; gomes@berkeley.edu).}%
\thanks{Corresponding author: Hendrik Chiche.}%
\thanks{A video demonstration of the pipeline is available at
\protect\url{https://youtu.be/mKtVg_gYwf8}.
Datasets and fine-tuned models are available at
\protect\url{https://huggingface.co/Chichonnade}.
Source code is available at
\protect\url{https://github.com/chichonnade/Vision-Based-Hand-Shadowing}.}%
\thanks{This is the author's accepted manuscript of an article accepted for
publication in IEEE Access. \textcopyright{}~2026 IEEE. Personal use of this
material is permitted. Permission from IEEE must be obtained for all other
uses, including reprinting, republishing, or redistribution.}}

\markboth{Chiche \textit{et al.}: Vision-Based Hand Shadowing for Robotic Manipulation via Inverse Kinematics}%
{Chiche \textit{et al.}: Vision-Based Hand Shadowing for Robotic Manipulation via Inverse Kinematics}

\maketitle

\begin{abstract}
Teleoperation of low-cost robotic manipulators remains challenging due to the
difficulty of retargeting human hand motion to robot joint commands.
We present an offline hand-shadowing inverse-kinematics (IK) retargeting
pipeline driven by a single egocentric RGB-D camera mounted on 3D-printed
glasses. The pipeline detects 21 hand landmarks per hand using MediaPipe
Hands, deprojects them into 3D via depth sensing, transforms them into the
robot coordinate frame, and solves a damped-least-squares IK problem to
produce joint commands for the SO-ARM101 robot (5 arm + 1 gripper joints).
A gripper controller maps thumb--index finger geometry to grasp aperture
with a multi-level fallback hierarchy. Actions are previewed in a physics
simulation before replay on the physical robot.
We evaluate the pipeline on a structured pick-and-place benchmark
(5-tile grid, 10 grasps per tile, 3 independent runs) achieving an
$86.7\% \pm 4.2\%$ success rate,
and compare it against four vision-language-action (VLA) policies
(ACT, SmolVLA, $\pi_{0.5}$, GR00T~N1.5) trained on
leader--follower teleoperation data.
We provide a quantitative error analysis of the pipeline, reporting
a mean IK position error of 36.4\,mm, trajectory smoothness metrics
showing 57--68\% jerk reduction from EMA smoothing, and an ablation
study over the smoothing parameter.
We also test the pipeline in unstructured real-world environments
(grocery store, pharmacy) and find that success is reduced to 9.3\%
due to hand occlusion by surrounding objects. To mitigate this, we
integrate WiLoR as an alternative hand detector, achieving an 8\%
improvement in hand detection rate over MediaPipe, highlighting both the
promise and current limitations of marker-free analytical retargeting.
\end{abstract}

\begin{IEEEkeywords}
Hand tracking, inverse kinematics, MediaPipe, PyBullet, robot teleoperation,
retargeting, SO-ARM101, vision-language-action models.
\end{IEEEkeywords}

% ============================= 1  INTRODUCTION ==============================
\section{Introduction}
\label{sec:introduction}
\IEEEPARstart{T}{eaching} robots to manipulate objects the way humans do is a
central goal of robotics research. Two dominant paradigms have emerged:
\emph{teleoperation}, where a human operator directly controls the robot
in real time, and \emph{imitation learning}, where a policy is trained
from recorded demonstrations~\cite{zhao2023act,cadena2025smolvla}.
Teleoperation provides immediate, interpretable control but traditionally
requires expensive hardware such as exoskeletons, leader--follower arm
pairs, or VR headsets~\cite{cheng2025opentv,ding2024bunny}. Imitation
learning can reduce the demonstration burden but demands careful data
collection, GPU-based training, and policy evaluation.

Recent marker-free teleoperation approaches such as
AnyTeleop~\cite{qin2023anyteleop} and
Dex-Cap~\cite{wang2024dexcap} demonstrate that vision-based hand
tracking can replace specialised hardware. However, these methods
typically require GPU-based pose estimators, multi-camera rigs, or
dexterous multi-finger hands. No prior work provides a complete,
CPU-only analytical retargeting pipeline from a single egocentric RGB-D
camera to a low-cost manipulator arm---requiring zero training data,
no GPU, and no specialised hardware beyond 3D-printed accessories.

This work studies how such an analytical inverse-kinematics (IK)
pipeline can bridge the teleoperation and imitation-learning paradigms
by converting first-person RGB-D recordings into robot trajectories
that can be replayed on hardware and reused as training data.
The pipeline is designed for \emph{offline} trajectory generation
rather than real-time control, which enables high-quality trajectory
processing without latency constraints.
Our contributions are:

\begin{enumerate}
  \item An end-to-end, CPU-only pipeline from egocentric RGB-D video
        to single-arm robot trajectory generation via damped-least-squares
        inverse kinematics, requiring no training data.
  \item A sim-to-real transfer workflow using PyBullet for trajectory
        preview and validation before physical deployment on the
        SO-ARM101 robot.
  \item A quantitative comparison of analytical IK retargeting with
        four VLA policies on a structured pick-and-place benchmark,
        with statistical confidence intervals across three independent
        runs.
  \item A comprehensive error analysis including IK solver accuracy
        (36.4\,mm mean position error), trajectory smoothness (57--68\%
        jerk reduction from EMA smoothing), and an EMA parameter
        ablation study.
  \item An in-the-wild evaluation in grocery store and pharmacy
        environments to assess robustness under clutter and occlusion.
\end{enumerate}

Across these studies, the IK pipeline reaches $86.7\% \pm 4.2\%$
success on the structured benchmark with zero training, while hand
occlusion emerges as the main limitation in unstructured scenes.

% ============================= 2  RELATED WORK ==============================
\section{Related Work}
\label{sec:related}

\subsection{Hand Pose Estimation}
Real-time hand tracking has advanced rapidly. MediaPipe
Hands~\cite{zhang2020mediapipe} runs on-device at 30\,Hz using a
two-stage BlazePalm + landmark pipeline predicting 21 keypoints,
and requires no GPU. We adopt MediaPipe for its CPU efficiency and
cross-platform availability.
WiLoR~\cite{potamias2024wilor} improves accuracy with a DarkNet
localiser followed by a ViT-based 3D reconstructor, achieving
state-of-the-art results on FreiHAND and HO3D benchmarks at over
130\,FPS but requires GPU inference. Both detectors output 21 keypoints
compatible with the MANO hand model topology~\cite{romero2017mano}.
Crucially, WiLoR's DarkNet-based hand localiser is more robust to
partial occlusion than MediaPipe's BlazePalm detector; we evaluate
WiLoR as an occlusion-mitigation strategy in Section~\ref{sec:occlusion}.

\subsection{Teleoperation Systems}
Open-TeleVision~\cite{cheng2025opentv} provides stereoscopic VR-based
teleoperation for humanoid robots. Bunny-VisionPro~\cite{ding2024bunny}
uses Apple Vision Pro for bimanual dexterous control with haptic
feedback. These systems achieve high fidelity but require specialised
VR hardware. Our approach uses only an RGB-D depth camera and
3D-printed accessories.

\subsection{Marker-Free Hand Retargeting}
AnyTeleop~\cite{qin2023anyteleop} provides a unified framework for
retargeting hand poses to diverse robot end-effectors using
GPU-accelerated optimisation.
Dex-Cap~\cite{wang2024dexcap} captures dexterous manipulation
demonstrations using a wearable motion-capture glove.
H2O~\cite{wang2024h2o} learns hand-to-object affordances for
retargeting. These approaches target dexterous multi-finger
hands or require GPU inference. Our pipeline is distinguished by its
simplicity: a single RGB-D camera, CPU-only MediaPipe detection,
analytical IK, and no training data.

\subsection{Imitation Learning}
ACT~\cite{zhao2023act} introduced action chunking with transformers for
fine-grained bimanual manipulation, achieving 80--90\% success with
10 minutes of demonstrations. SmolVLA~\cite{cadena2025smolvla} is a
450M-parameter vision-language-action model trainable on a single GPU.
$\pi_0$~\cite{black2024pi0} is a generalist robot foundation model
using flow matching over a VLM backbone, trained across 7 robot
embodiments and 68 tasks.

\subsection{Low-Cost Robotics}
The SO-ARM100/101 is a 6-DOF arm using STS3215 bus servos
with 30\,kg$\cdot$cm torque~\cite{soarm100}. LeRobot~\cite{lerobot2024}
provides a hardware-agnostic Python framework that standardises data
collection, training, and deployment across robot platforms.

\subsection{Physics Simulation}
PyBullet~\cite{coumans2021pybullet} provides real-time rigid body
dynamics with built-in inverse kinematics via the Bullet Physics SDK.
It supports URDF loading, position/velocity/torque control, and
GPU-accelerated rendering, making it suitable for rapid prototyping
and sim-to-real transfer.

% ============================= 3  PIPELINE OVERVIEW =========================
\section{Pipeline Overview}
\label{sec:pipeline}

We refer to the complete processing chain---from egocentric RGB-D
capture through hand detection, 3D deprojection, coordinate
transformation, and inverse-kinematics solving to robot joint
commands---as the \emph{IK retargeting pipeline}
(Fig.~\ref{fig:architecture}). It requires an Intel RealSense
D400 camera~\cite{keselman2017realsense} mounted on 3D-printed glasses
and a single SO-ARM101 follower arm.

Fig.~\ref{fig:architecture} shows the overall data flow through the
pipeline stages.

\begin{figure}[t]
\centering
\begin{tikzpicture}[
  node distance=0.55cm,
  stage/.style={rectangle, draw, rounded corners=2pt,
    minimum width=5.0cm, minimum height=0.55cm,
    align=center, font=\footnotesize},
  data/.style={font=\scriptsize, align=center},
  output/.style={rectangle, draw, rounded corners=2pt,
    minimum width=2.1cm, minimum height=0.5cm,
    align=center, font=\footnotesize},
  arr/.style={-{Stealth[length=4pt]}, thick}
]
\node[stage] (cam)   {(A) RGB-D Camera};
\node[data, below=0.2cm of cam] (dcam)
  {\itshape RGB image + depth map};
\node[stage, below=0.2cm of dcam] (hand)
  {(B) Hand Landmark Detection (MediaPipe)};
\node[data, below=0.2cm of hand] (dhand)
  {\itshape $\{(u_i, v_i)\}_{i=0}^{20}$ --- 21 landmarks};
\node[stage, below=0.2cm of dhand] (depth)
  {(C) Depth Deprojection};
\node[data, below=0.2cm of depth] (ddepth)
  {\itshape $\vect{P}_{\text{cam}}$ --- 3D camera coordinates};
\node[stage, below=0.2cm of ddepth] (coord)
  {(D) Coordinate Transform};
\node[data, below=0.2cm of coord] (dcoord)
  {\itshape $\vect{P}_{\text{robot}}$ --- 3D robot-frame coordinates};
\node[stage, below=0.2cm of dcoord] (tgt)
  {Target Pose Computation};
\node[data, below=0.2cm of tgt] (dtgt)
  {\itshape $(\vect{p}_{\text{target}}, \vect{q}_{\text{target}})$ in robot frame};
\node[stage, below=0.2cm of dtgt] (ik)
  {(E) IK Solver + Gripper Controller};
\node[data, below=0.2cm of ik] (dik)
  {\itshape $\vect{q}$ --- joint angles};
\node[output, below=0.45cm of dik, xshift=-1.25cm] (sim)   {(F) Sim Preview};
\node[output, below=0.45cm of dik, xshift=1.25cm]  (robot) {(G) SO-ARM101};
\draw[arr] (cam)   -- (dcam);
\draw[arr] (dcam)  -- (hand);
\draw[arr] (hand)  -- (dhand);
\draw[arr] (dhand) -- (depth);
\draw[arr] (depth) -- (ddepth);
\draw[arr] (ddepth)-- (coord);
\draw[arr] (coord) -- (dcoord);
\draw[arr] (dcoord)-- (tgt);
\draw[arr] (tgt)   -- (dtgt);
\draw[arr] (dtgt)  -- (ik);
\draw[arr] (ik)    -- (dik);
\draw[arr] (dik.south) -- ++(0,-0.1) -| (sim.north);
\draw[arr] (dik.south) -- ++(0,-0.1) -| (robot.north);
\end{tikzpicture}
\caption{\textbf{Pipeline architecture.}
  Each stage (A--E) transforms the data from egocentric RGB-D frames
  to robot joint commands. Intermediate representations are shown
  between stages. The output feeds both a simulation preview (F)
  and the physical robot (G).}
\label{fig:architecture}
\end{figure}

\subsection{Hardware}
\label{sec:hardware}

\begin{figure}[t]
\centering
\begin{minipage}[b]{0.35\columnwidth}
\centering
\includegraphics[height=5cm]{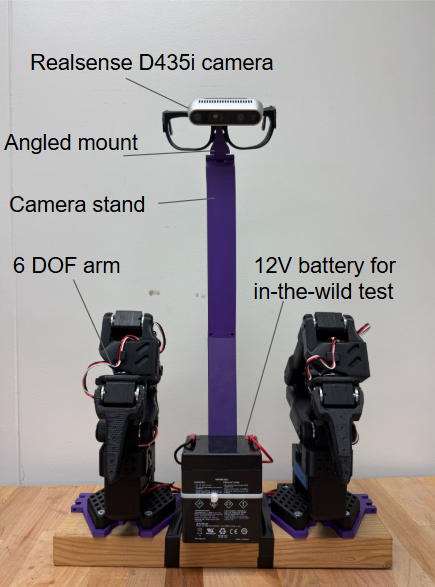}
\end{minipage}%
\hspace{0.05\columnwidth}%
\begin{minipage}[b]{0.55\columnwidth}
\centering
\begin{tikzpicture}[scale=0.5, font=\footnotesize]
  % Ground line
  \draw[thick, gray] (-1.5,0) -- (3.0,0);
  % Robot base
  \draw[thick, fill=gray!20] (-0.5,0) rectangle (0.5,0.6);
  \node[below] at (0,-0.2) {\scriptsize Robot base};
  % Robot coordinate system at base
  \draw[-{Stealth[length=3pt]}, thick, green!60!black] (0,0.3) -- (0.9,0.3)
    node[right] {$y_r$};
  \draw[-{Stealth[length=3pt]}, thick, blue] (0,0.3) -- (0,1.2)
    node[left] {$z_r$};
  % Camera stand
  \draw[thick] (0,0.6) -- (0,8.0);
  % Camera body (rotated by theta around top of stand)
  \draw[thick, fill=blue!20, rotate around={-50:(0,8.0)}]
    (-0.6,8.0) rectangle (0.6,8.5);
  \node[right] at (0.8,8.3) {\scriptsize Camera};
  % ty arrow at the top (forward offset from stand to camera origin)
  \pgfmathsetmacro{\camx}{0.6*sin(50)}
  \pgfmathsetmacro{\camy}{8.3-0.6*cos(50)}
  \draw[<->, thick] (0,8.7) -- (1.2,8.7);
  \node[above, font=\scriptsize] at (0.6,8.7) {$t_y{=}0.049$\,m};
  % Horizontal reference at camera level (for theta arc)
  \draw[->, thin, gray] (0,8.0) -- (2.2,8.0);
  % Camera coordinate triad at the camera lens position
  % z_c along optical axis
  \draw[-{Stealth[length=3pt]}, thick, blue]
    (\camx,\camy) -- ({\camx+1.1*cos(50)},{\camy-1.1*sin(50)})
    node[below right, font=\tiny] {$z_c$};
  % x_c perpendicular to z_c in tilt plane
  \draw[-{Stealth[length=3pt]}, thick, red]
    (\camx,\camy) -- ({\camx-0.8*sin(50)},{\camy-0.8*cos(50)})
    node[below left, font=\tiny] {$x_c$};
  % Optical axis (dashed extension of z_c)
  \draw[-{Stealth[length=4pt]}, thick, dashed, blue!50]
    (\camx,\camy) -- ({\camx+2.2*cos(50)},{\camy-2.2*sin(50)});
  % Angle theta arc (from horizontal to z_c)
  \draw[thick] (0.6,8.0) arc (0:-50:0.6);
  \node[font=\scriptsize] at (0.8,7.6) {$\theta$};
  % Height annotation (tz)
  \draw[<->, thick] (2.5,0) -- (2.5,8.0);
  \node[right, font=\scriptsize] at (2.5,4.0) {$t_z{=}0.48$\,m};
\end{tikzpicture}
\end{minipage}
\caption{\textbf{Hardware setup.} Left: annotated photo showing the
  RealSense D435i camera, angled mount, camera stand, 6-DOF arm, and
  12V battery for in-the-wild deployment.
  Right: schematic showing the camera tilt angle
  $\theta = 50^{\circ}$ below horizontal and height offset
  $t_z = 0.48$\,m relative to the robot base
  (see Section~\ref{sec:transform}).}
\label{fig:bench}
\end{figure}

The egocentric sensor platform consists of an Intel RealSense D400-series
stereo depth camera mounted on 3D-printed glasses using three PLA/ABS
parts (frame, two temple branches, and a camera mount bracket), secured
with M1.5, M2, and M3 screws. The camera connects via USB-C 3.1 Gen~1
and streams synchronised RGB and depth at 640$\times$480 resolution and
30\,FPS. The robot platform uses a single SO-ARM101 follower arm
with 6 revolute joints (5 arm + 1 gripper) driven by STS3215
Feetech bus servos. The complete lab bench setup, including the robot
and egocentric camera stand, is shown in Fig.~\ref{fig:bench}.

% ============================= 4  METHODS ===================================
\section{Methods}
\label{sec:methods}

\subsection{RGB-D Capture and Camera Model}
\label{sec:camera}

The RealSense camera provides aligned RGB-D frames via the
RealSense SDK. We denote a 3D point in camera coordinates as
$\vect{P}_{\text{cam}} = (X, Y, Z)^\top$ and model the camera
using the pinhole projection:
\begin{equation}\label{eq:projection}
  \begin{bmatrix} u \\ v \end{bmatrix}
  = \begin{bmatrix} f_x & 0 & c_x \\ 0 & f_y & c_y \end{bmatrix}
    \begin{bmatrix} X/Z \\ Y/Z \\ 1 \end{bmatrix},
\end{equation}
where $(f_x, f_y)$ are focal lengths and $(c_x, c_y)$ is the principal
point, all extracted automatically from the camera stream at
initialisation. The camera supports recording to \texttt{.bag} files for
offline processing and to MP4 for archival.

\subsection{Hand Pose Estimation}
\label{sec:handpose}

We use MediaPipe Hands~\cite{zhang2020mediapipe} for 2D hand detection
and landmark localisation. MediaPipe's two-stage architecture---a
BlazePalm detector that locates hands via oriented bounding boxes,
followed by a lightweight landmark regression network---predicts
21 keypoints per hand in real time on CPU. The detector outputs
left/right hand classification and 2D keypoint coordinates
$\{(u_i, v_i)\}_{i=0}^{20}$ covering the wrist, thumb, index, middle,
ring, and pinky finger joints.

Let $\vect{P}_t^{\text{raw}} = \{(u_i^{\text{raw}}, v_i^{\text{raw}})\}_{i=0}^{20}$
denote the raw MediaPipe landmark array at time $t$.
To reduce temporal jitter, we apply exponential moving average (EMA)
smoothing:
\begin{equation}\label{eq:ema}
  \vect{P}_{t} = \alpha\, \vect{P}_t^{\text{raw}}
                + (1 - \alpha)\, \vect{P}_{t-1},
  \quad \alpha = 0.8,
\end{equation}
where $\vect{P}_{t}$ is the smoothed landmark array.
The smoothed per-landmark coordinates $(u_i, v_i)$ are
then used in subsequent pipeline stages.

\subsection{Depth-Based 3D Reconstruction}
\label{sec:depth3d}

We define two key gripper-reference landmarks: $\vect{P}_1$ (thumb
metacarpophalangeal joint, MCP) and $\vect{P}_2$ (index finger MCP),
following the MediaPipe 21-keypoint hand topology.
Each 2D landmark $(u_i, v_i)$ is deprojected to a 3D camera-space point
$\vect{P}_{\text{cam}}$ using the depth image $D$ (in metres) and camera
intrinsics:
\begin{equation}\label{eq:deproject}
  \vect{P}_{\text{cam}}
  = D[u_i, v_i]
    \begin{bmatrix}
      (u_i - c_x) / f_x \\
      (v_i - c_y) / f_y \\
      1
    \end{bmatrix},
\end{equation}
where $D[u_i, v_i]$ is the depth at pixel $(u_i, v_i)$ in metres.
Depth values outside the valid range of 0.1--5.0\,m are marked invalid.

A \emph{depth fallback} mechanism handles the critical case where
exactly one of the two gripper-defining landmarks ($\vect{P}_1$ or
$\vect{P}_2$) has invalid depth: the valid landmark's depth is
substituted for the failed one, since these adjacent landmarks are
typically at similar depths. A hand pose is rejected entirely if fewer
than 50\% of its 21 landmarks have valid depth.

\subsection{Camera-to-Robot Coordinate Transform}
\label{sec:transform}

All 21 landmarks in camera coordinates $\vect{P}_{\text{cam}}$ are
transformed to the robot base frame by:
\begin{equation}\label{eq:transform}
  \vect{P}_{\text{robot}} = \mat{R}\,\vect{P}_{\text{cam}} + \vect{t}.
\end{equation}
All landmark symbols used in subsequent sections ($\vect{P}_1$,
$\vect{P}_2$, fingertip positions, etc.) refer to robot-frame
coordinates after this transform has been applied.

The rotation matrix $\mat{R}$ accounts for the
camera tilt angle $\theta = 50^{\circ}$ below the horizontal
(Fig.~\ref{fig:bench}):
\begin{equation}\label{eq:Rcam}
  \mat{R} =
  \begin{bmatrix}
    -1 & 0 & 0 \\
    0 & \sin\theta & -\cos\theta \\
    0 & -\cos\theta & -\sin\theta
  \end{bmatrix},
\end{equation}
with translation $\vect{t} = (0.04,\; -0.049,\; 0.48)^\top$
metres. These parameters---including the $50^{\circ}$ mounting angle and
translation offsets---are extracted directly from the SolidWorks CAD
assembly that models both the robot and the camera mount bracket at
their exact physical positions.
The $x$-axis negation mirrors the image horizontally (the camera
observes hands from a first-person perspective, while the robot faces
the operator), and the rotation about the $x$-axis maps the downward
camera view to the robot's forward--up coordinate system.

\subsection{Target Pose Computation}
\label{sec:targetpose}

The target end-effector pose $(\vect{p}_{\text{target}},
\vect{q}_{\text{target}})$ consists of a position and an orientation
quaternion, both expressed in the robot base frame. The target position
is the midpoint of the two gripper-defining landmarks:
\begin{equation}\label{eq:targetpos}
  \vect{p}_{\text{target}} = \frac{1}{2}
    \bigl(\vect{P}_1 + \vect{P}_2\bigr).
\end{equation}

The target orientation $\vect{q}_{\text{target}}$ is derived from
three orthogonal axes constructed from hand geometry
(Fig.~\ref{fig:gripper_vectors}):
\begin{align}
  \vect{e}_1 &= \frac{\vect{P}_2 - \vect{P}_1}
    {\|\vect{P}_2 - \vect{P}_1\|},
    \label{eq:e1}
\end{align}
which represents the gripper width direction. The average finger
pointing direction $\vect{d}$ is computed from the unit vectors
$\vect{u}_{\text{thumb}}$ and $\vect{u}_{\text{index}}$ pointing from
each MCP landmark toward its respective fingertip:
\begin{align}
  \vect{d} &= \frac{\vect{u}_{\text{thumb}} +
    \vect{u}_{\text{index}}}
    {\|\vect{u}_{\text{thumb}} + \vect{u}_{\text{index}}\|}.
    \label{eq:davg}
\end{align}
We average the \emph{unit} vectors rather than the raw finger
direction vectors because normalising before averaging ensures that
both fingers contribute equally to the pointing direction regardless
of their measured 3D lengths, which can differ due to depth noise.
The remaining orthonormal axes are obtained via cross products:
\begin{align}
  \vect{e}_3 &= \frac{\vect{e}_1 \times \vect{d}}
    {\|\vect{e}_1 \times \vect{d}\|}, \label{eq:e3} \\
  \vect{e}_2 &= \vect{e}_3 \times \vect{e}_1. \label{eq:e2}
\end{align}
The target orientation quaternion $\vect{q}_{\text{target}}$ is the
quaternion formed by $[\vect{e}_1 \;\; \vect{e}_2 \;\; \vect{e}_3]$.

When fingertip landmarks are unavailable (due to occlusion or depth
failure), a \emph{fallback orientation} is computed using only
$\vect{P}_1$, $\vect{P}_2$, and the wrist landmark, replacing
$\vect{d}$ with the wrist-to-gripper-centre vector.

\begin{figure}[t]
\centering
\begin{tikzpicture}[scale=0.9, font=\footnotesize,
  vec/.style={-{Stealth[length=3pt]}, thick}]
  % P1 (thumb MCP) on the left, P2 (index MCP) on the right
  \coordinate (tmcp) at (0,0);
  \coordinate (imcp) at (4,0);
  \coordinate (target) at (2,0);
  % Fingertips at different angles
  % u_thumb: tilted inward (toward centre) at ~70 deg from horizontal
  \coordinate (ttip) at ({0+2.2*cos(70)},{2.2*sin(70)});
  % u_index: tilted inward at ~110 deg from horizontal (toward centre)
  \coordinate (itip) at ({4+2.2*cos(110)},{2.2*sin(110)});
  % d = normalised average of u_thumb and u_index unit vectors
  % u_thumb direction: (cos70, sin70) = (0.342, 0.940)
  % u_index direction: (cos110, sin110) = (-0.342, 0.940)
  % average: (0, 0.940), normalised: (0, 1) -> straight up
  % But to make it visually interesting, use slightly asymmetric angles:
  % u_thumb at 80 deg, u_index at 100 deg
  % u_thumb dir: (cos80, sin80) = (0.174, 0.985)
  % u_index dir: (cos100, sin100) = (-0.174, 0.985)
  % avg: (0, 0.985) -> nearly vertical, normalised: (0, 1)
  % Let's use 75 and 105 for more visible difference:
  % u_thumb: (cos75, sin75) = (0.259, 0.966)
  % u_index: (cos105, sin105) = (-0.259, 0.966)
  % avg: (0, 0.966) -> vertical
  % Use asymmetric: u_thumb at 80, u_index at 60 (from horizontal)
  % u_thumb dir: (cos80, sin80) = (0.174, 0.985)
  % u_index dir: (cos120, sin120) = (-0.5, 0.866)
  % avg: (-0.163, 0.925), normalised ~ (-0.174, 0.985)
  % d points slightly left and up
  \pgfmathsetmacro{\uthx}{cos(80)}
  \pgfmathsetmacro{\uthy}{sin(80)}
  \pgfmathsetmacro{\uix}{cos(120)}
  \pgfmathsetmacro{\uiy}{sin(120)}
  \pgfmathsetmacro{\davgx}{\uthx+\uix}
  \pgfmathsetmacro{\davgy}{\uthy+\uiy}
  \pgfmathsetmacro{\dnorm}{sqrt(\davgx*\davgx+\davgy*\davgy)}
  \pgfmathsetmacro{\dx}{\davgx/\dnorm}
  \pgfmathsetmacro{\dy}{\davgy/\dnorm}
  % Redefine fingertip positions using these angles (same length as d)
  \def\arrlen{1.8}
  \coordinate (ttip) at ({0+\arrlen*\uthx},{\arrlen*\uthy});
  \coordinate (itip) at ({4+\arrlen*\uix},{\arrlen*\uiy});
  % e1: horizontal, gripper width
  \draw[vec, blue, very thick] (tmcp) -- (imcp)
    node[midway, below=12pt] {$\vect{e}_1$};
  % Finger unit vectors at different angles (same visual length)
  \draw[vec, magenta, dashed] (tmcp) -- (ttip)
    node[midway, left] {$\vect{u}_{\text{thumb}}$};
  \draw[vec, magenta, dashed] (imcp) -- (itip)
    node[midway, right] {$\vect{u}_{\text{index}}$};
  % d: normalised average, from target (same visual length)
  \draw[vec, magenta, very thick] (target) -- ++({\dx*\arrlen},{\dy*\arrlen})
    node[right=2pt] {$\vect{d}$};
  % Points
  \fill (tmcp) circle (2.5pt) node[below left] {$\vect{P}_1$};
  \fill (imcp) circle (2.5pt) node[below right] {$\vect{P}_2$};
  \fill[red] (target) circle (3pt)
    node[below=4pt] {$\vect{p}_{\text{target}}$};
  \fill[gray] (ttip) circle (2pt) node[above] {tip};
  \fill[gray] (itip) circle (2pt) node[above] {tip};
  % Equations on the right
  \node[align=left, font=\scriptsize] at (6.5, 1.25)
    {$\vect{e}_3 = \vect{e}_1 \times \vect{d}$\\[2pt]
     $\vect{e}_2 = \vect{e}_3 \times \vect{e}_1$\\[2pt]
     ($\vect{e}_3$ out of page)};
\end{tikzpicture}
\caption{\textbf{Gripper orientation vectors.}
  $\vect{e}_1$ is the normalised gripper width direction
  ($\vect{P}_1 \to \vect{P}_2$).
  $\vect{u}_{\text{thumb}}$ and $\vect{u}_{\text{index}}$ are unit
  vectors from each MCP to its fingertip; their normalised average
  $\vect{d}$ (Eq.~\ref{eq:davg}) serves as a reference pointing
  direction. Note that $\vect{d}$ is generally \emph{not} orthogonal
  to $\vect{e}_1$; the orthonormal frame is obtained by computing
  $\vect{e}_3 = \vect{e}_1 \times \vect{d}$ (normalised, out of page)
  and $\vect{e}_2 = \vect{e}_3 \times \vect{e}_1$
  (Eqs.~\ref{eq:e3}--\ref{eq:e2}).}
\label{fig:gripper_vectors}
\end{figure}

\subsection{Inverse Kinematics}
\label{sec:ik}

Given the target pose $(\vect{p}_{\text{target}},
\vect{q}_{\text{target}})$, we solve for joint angles
$\vect{q} = (q_1, \ldots, q_5)$ of the 5 arm joints
by solving the following damped-least-squares (DLS)
problem~\cite{wampler1986dls}:
\begin{equation}\label{eq:ik}
  \vect{q}^* = \arg\min_{\vect{q}}
  \bigl\|
    \text{FK}(\vect{q}) - (\vect{p}_{\text{target}},
                                  \vect{q}_{\text{target}})
  \bigr\|^2
  + \lambda \|\vect{q} - \vect{q}_{\text{rest}}\|^2,
\end{equation}
where $\text{FK}(\cdot)$ is the forward kinematics function,
$\vect{q}_{\text{rest}}$ is the current joint state (used as the
rest pose), and $\lambda$ captures per-joint damping extracted from the
URDF. The solver uses the PyBullet IK engine and runs up to 100
iterations with a residual threshold of $10^{-4}$.

Because the arm has only 5 revolute joints for positioning, it cannot
independently satisfy a full 6-DOF target pose (3 position +
3 orientation). The DLS formulation finds the best compromise,
prioritising position accuracy over orientation---an analysis of this
trade-off is presented in Section~\ref{sec:ikaccuracy}.

A second EMA filter ($\alpha_{\text{IK}} = 0.5$) is applied to the IK
solution to suppress temporal jitter in joint space:
\begin{equation}
  \hat{\vect{q}}_t = \alpha_{\text{IK}}\,\vect{q}^*_t
                        + (1 - \alpha_{\text{IK}})\,\hat{\vect{q}}_{t-1}.
\end{equation}
This is distinct from the landmark-space EMA
($\alpha_{\text{lm}} = 0.8$, Eq.~\ref{eq:ema}), which smooths 2D
keypoints before deprojection.

A safety check rejects any target with $z < 0.05$\,m to prevent
ground-plane collisions.

\subsection{Gripper Control}
\label{sec:gripper}

Let $\vect{p}_{\text{thumb}}$ and $\vect{p}_{\text{index}}$ denote
the thumb tip and index fingertip positions in robot-frame coordinates.
The target gripper angle $\phi_{\text{target}}$ is computed from the
angular separation between vectors from the gripper centre
($\vect{p}_{\text{target}}$) to these two fingertips:
\begin{equation}\label{eq:gripper}
  \phi_{\text{target}} = \arccos\!\left(
    \frac{(\vect{p}_{\text{thumb}} - \vect{p}_{\text{target}})
          \cdot
          (\vect{p}_{\text{index}} - \vect{p}_{\text{target}})}
         {\|\vect{p}_{\text{thumb}} - \vect{p}_{\text{target}}\|\;
          \|\vect{p}_{\text{index}} - \vect{p}_{\text{target}}\|}
  \right),
\end{equation}
clamped to the acute range $[0, \pi/2]$ and then to the gripper joint
limits $[\phi_{\min}, \phi_{\max}] = [0.087, 1.658]$\,rad after
applying an offset of $-0.175$\,rad for tighter grip calibration.

A multi-level depth fallback mechanism handles the critical case where
exactly one of the two fingertip landmarks has invalid depth.
When the primary landmarks (thumb tip and index fingertip) are
unavailable, the corresponding knuckle landmarks are used instead;
when those also fail, the last valid gripper angle is held via temporal
persistence; as a final default, the gripper is set to its mid-open
position $(\phi_{\min}+\phi_{\max})/2$. This graceful degradation
strategy ensures that a single missing landmark does not cause
catastrophic gripper behaviour.

\subsection{Simulation Preview}
\label{sec:simulation}

Before deploying on the physical robot, all IK-generated trajectories
are previewed in a PyBullet simulation environment. The simulation
loads the SO-ARM101 URDF with 7 joints per arm
(1 fixed base + 5 revolute arm + 1 revolute gripper); only the
right arm is used in our experiments. The simulation
runs at 240\,Hz. Position-control PID parameters were empirically
tuned to match the motion speed and response characteristics of the
physical robot arm (see Section~\ref{sec:simreal} for a quantitative
comparison of simulated and physical tracking behaviour).

\begin{figure}[t]
\centering
\includegraphics[width=0.9\columnwidth]{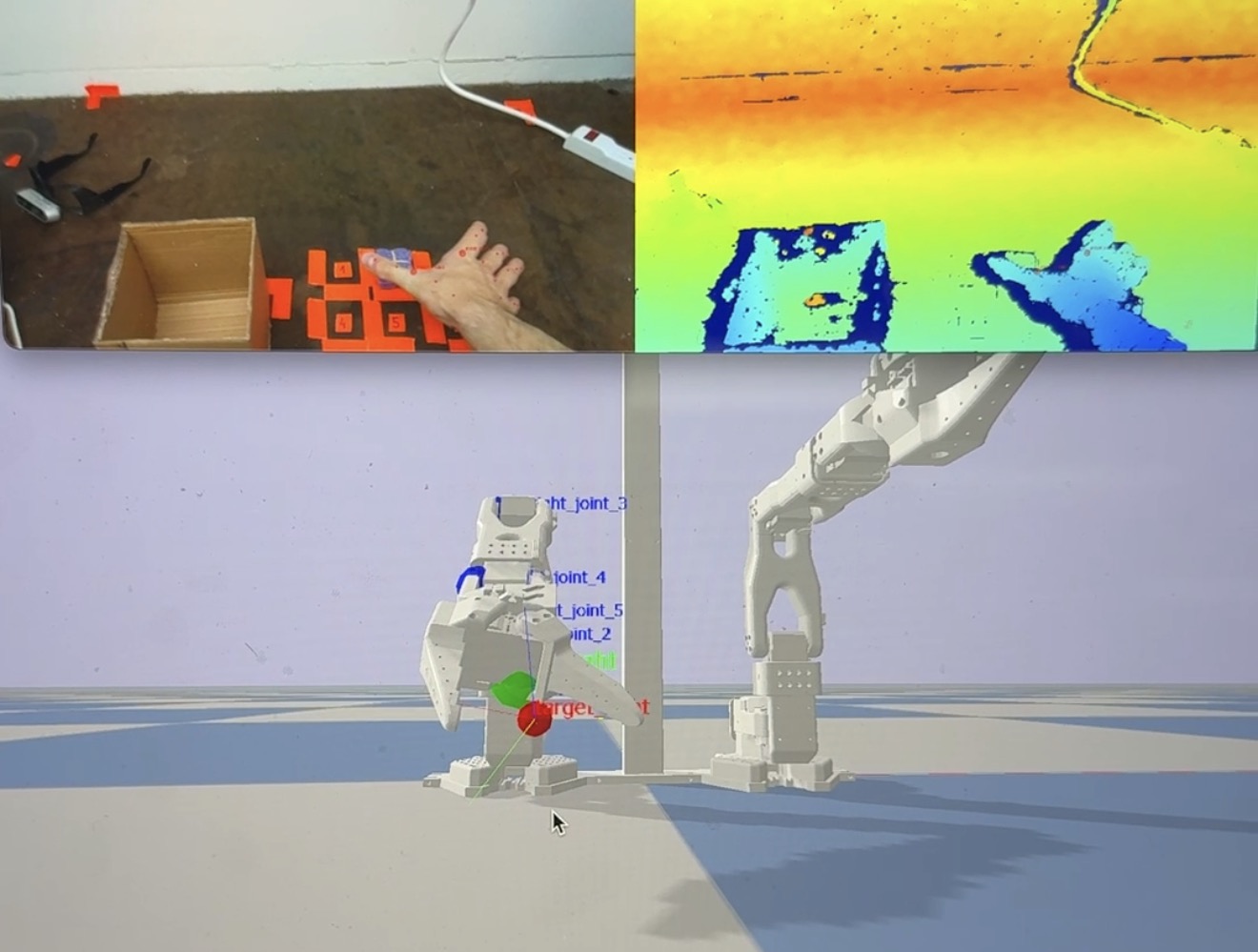}
\caption{\textbf{PyBullet simulation preview.} Top-left: RGB frame from the
  egocentric camera showing the operator's hand. Top-right: depth
  colour map. Bottom: robot arm in PyBullet with debug joint
  labels and IK target markers (green/red spheres), tracking the
  hand-derived trajectory.}
\label{fig:simulation}
\end{figure}

Fig.~\ref{fig:simulation} shows the simulation preview interface.
The top-left panel displays the RGB frame from the egocentric
camera with the operator's hand visible; the top-right panel shows the
corresponding depth map colour-coded by distance; and the bottom panel
renders the robot in PyBullet tracking the hand-derived IK
targets. This three-panel view allows the operator to
verify that the computed joint trajectories faithfully reproduce the
intended hand motion before committing to physical execution.
Beyond trajectory verification, the simulation also exports
demonstration data (\texttt{.npy} action files) that can be used
to train imitation-learning policies.

\subsection{Physical Deployment}
\label{sec:deployment}

Validated actions are deployed on the physical SO-ARM101 via the
LeRobot~\cite{lerobot2024} framework. Each solved joint angle $q_i$
is linearly mapped to a normalised action
$a_{\text{norm}} = (q_i - q_{\min}) / (q_{\max} - q_{\min}) \in [0, 1]$
using the per-joint limits in Table~\ref{tab:joints}, then converted
to a motor command $a_{\text{motor}}$ following the LeRobot
motor-space convention for STS3215 servos:
\begin{equation}
  a_{\text{motor}} =
  \begin{cases}
    (a_{\text{norm}} - 0.5) \times 200 & \text{arm joints} \;\in [-100, 100], \\
    a_{\text{norm}} \times 100 & \text{gripper} \;\in [0, 100].
  \end{cases}
\end{equation}

Servo PID coefficients are configured for smooth motion (Table~\ref{tab:servo}),
and acceleration/velocity limits prevent abrupt movements. Three gripper
modes are supported: \emph{Normal} (raw angle pass-through),
\emph{Binary} (threshold at $60^{\circ}$; outputs fully open or closed),
and \emph{Offset} (adds a configurable offset for tighter grasps).

\begin{figure}[t]
\centering
\includegraphics[width=0.9\columnwidth]{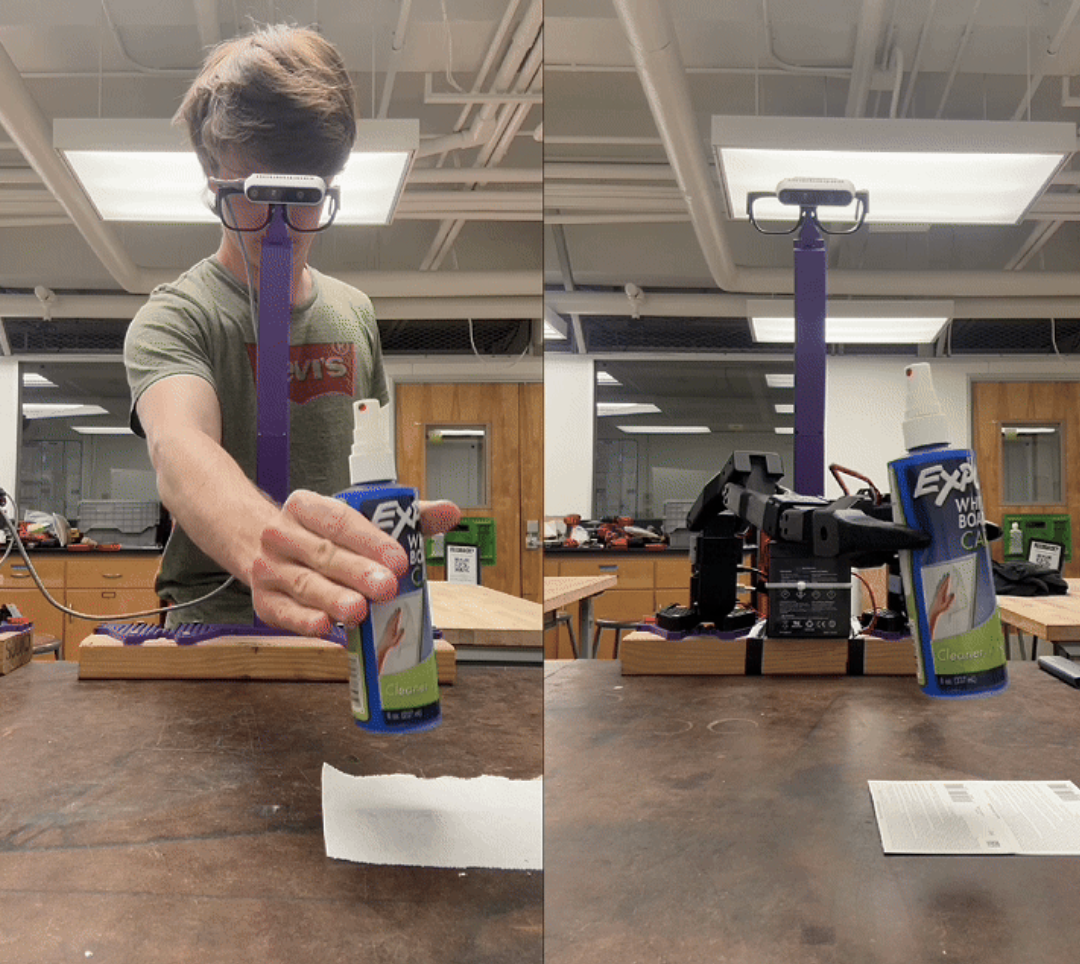}
\caption{\textbf{Hand shadowing.} Left: the operator wearing the
  RealSense glasses performs a grasp. Right: the SO-ARM101 robot
  mirrors the hand pose through the IK pipeline.}
\label{fig:shadowing}
\end{figure}

Fig.~\ref{fig:shadowing} shows the hand-shadowing result:
the operator (left) grasps an object while wearing the camera glasses,
and the robot (right) mirrors the grasp posture via the
IK pipeline after offline processing.

% ============================= 5  EXPERIMENTAL SETUP ========================
\section{Experimental Setup}
\label{sec:experiments}

\subsection{Hardware and Software}
\begin{itemize}
  \item \textbf{Camera}: Intel RealSense D400, 640$\times$480 RGB+D at 30\,FPS.
  \item \textbf{Robot}: Single SO-ARM101 follower arm (STS3215 servos,
        30\,kg$\cdot$cm torque).
  \item \textbf{Glasses}: Custom 3D-printed PLA mount (3 parts: frame,
        2$\times$ branch, mount bracket).
  \item \textbf{Compute}: Apple M-series or x86 laptop (MediaPipe runs
        on CPU; the IK solver requires no GPU).
\end{itemize}

Table~\ref{tab:joints} lists the joint angle ranges that define the
normalisation bounds for physical deployment and the IK solver's
feasible set. Table~\ref{tab:servo} gives the servo PID and
motion-limit parameters.

\begin{table}[t]
\caption{\textbf{SO-ARM101 Joint Angle Ranges (Radians)}}
\label{tab:joints}
\centering
\begin{tabular}{lrrr}
\hline
\textbf{Joint} & \textbf{Min} & \textbf{Max} & \textbf{Range ($^{\circ}$)} \\
\hline
Shoulder pan   & $-1.920$ & $+1.920$ & $220.0$ \\
Shoulder lift  & $-1.745$ & $+1.745$ & $200.0$ \\
Elbow flex     & $-1.745$ & $+1.571$ & $190.0$ \\
Wrist flex     & $-1.658$ & $+1.658$ & $190.0$ \\
Wrist roll     & $-2.793$ & $+2.793$ & $320.0$ \\
Gripper        & $-0.175$ & $+1.745$ & $110.0$ \\
\hline
\end{tabular}
\end{table}

\begin{table}[t]
\caption{\textbf{Physical Robot Servo Parameters}}
\label{tab:servo}
\centering
\begin{tabular}{lr}
\hline
\textbf{Parameter} & \textbf{Value} \\
\hline
P coefficient               & 12 \\
I coefficient               & 0 \\
D coefficient               & 24 \\
Arm acceleration limit      & 50 / 254 \\
Gripper acceleration limit  & 100 / 254 \\
Arm velocity limit          & 1500\,RPM \\
Gripper velocity limit      & 3000\,RPM \\
\hline
\end{tabular}
\end{table}

\subsection{Task and Protocol}
\label{sec:task}

\begin{figure}[t]
\centering
\begin{minipage}[b]{0.45\columnwidth}
\centering
\begin{tikzpicture}[scale=0.5, font=\footnotesize]
  % Grid
  \foreach \r/\y in {1/4, 4/2, 7/0} {
    \foreach \c in {0,1,2} {
      \pgfmathtruncatemacro{\tile}{\r + \c}
      \draw (\c*1.5, \y) rectangle ++(1.2, 1.2);
      \node at (\c*1.5+0.6, \y+0.6) {\tile};
    }
  }
  % Box
  \draw[thick, fill=gray!15] (-2.2, 1.5) rectangle (-0.5, 3.5);
  \node at (-1.35, 2.5) {\tiny Box};
  % Robot
  \draw[thick, fill=gray!30] (0.8, -1.2) rectangle (2.0, -0.5);
  \node at (1.4, -0.85) {\tiny Robot};
\end{tikzpicture}
\end{minipage}%
\hfill
\begin{minipage}[b]{0.52\columnwidth}
\centering
\includegraphics[width=\textwidth]{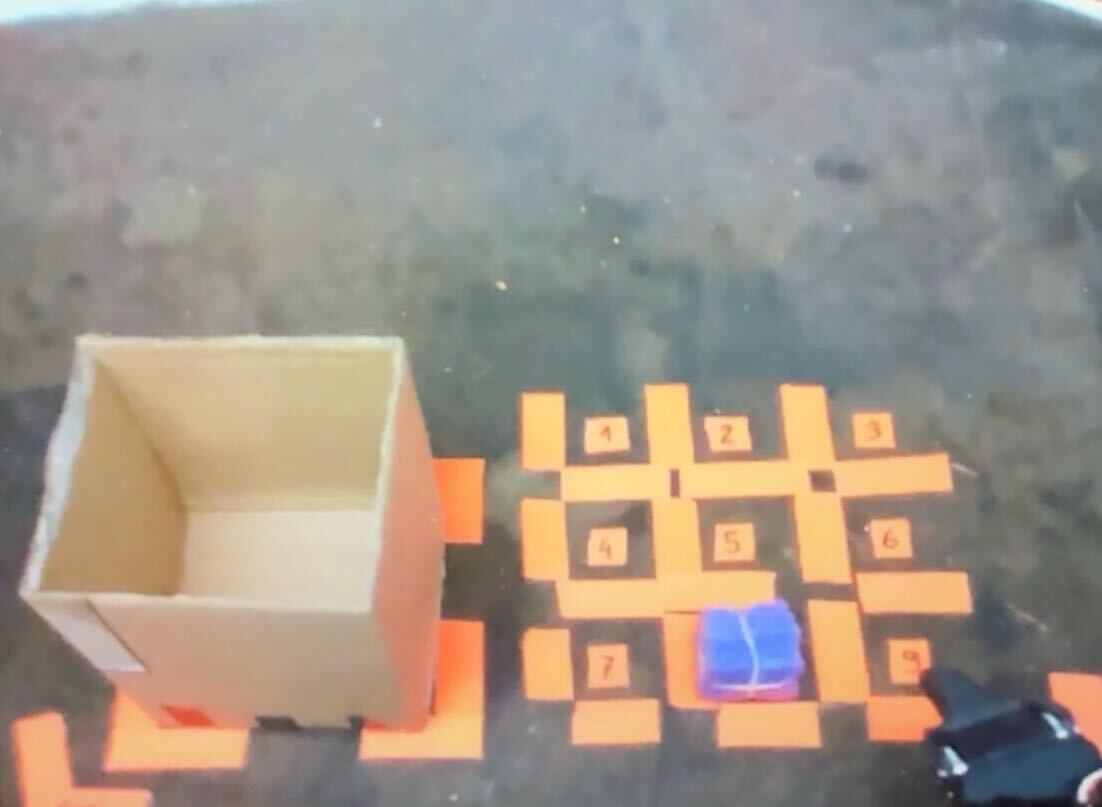}
\end{minipage}
\caption{\textbf{Pick-and-place benchmark layout.} Left: $3 \times 3$
  tile grid numbered \#1--\#9 from the robot's perspective, with the
  drop box to the left. Right: camera point-of-view showing the grid
  with the purple cube on tile~\#5. Tiles \#1--\#5 are used for
  evaluation; \#6--\#9 are excluded due to hand self-occlusion.}
\label{fig:grabbing}
\end{figure}

The benchmark task is: \emph{``Grab the purple cube and drop it in the
box.''} The cube is a soft foam block ($5 \times 5 \times 5$\,cm) placed
on a $3 \times 3$ tile grid (tiles numbered \#1--\#9), with a
$25 \times 25$\,cm cardboard box positioned to the left of the grid.
Fig.~\ref{fig:grabbing} shows the task layout.

This pick-and-place task was deliberately chosen as the benchmark
because it is the simplest manipulation task that exercises the
\emph{entire} pipeline end-to-end---from hand tracking through IK
solving to physical grasping---while remaining reliably achievable
within the hardware constraints of the SO-ARM101 (5 arm joints,
a two-finger gripper, and ${\sim}$300\,mm reach). More complex tasks
(e.g., in-hand reorientation, multi-object sorting) would conflate
pipeline failures with fundamental hardware limitations, preventing
a clean evaluation of the retargeting approach itself.

For the IK pipeline, the operator wears the glasses and
performs the task naturally. The pipeline records 30\,FPS RGB-D to a
\texttt{.bag} file, processes it offline to extract joint-angle actions,
previews the trajectory in PyBullet (Section~\ref{sec:simulation}),
and then deploys on the physical robot.

The reported benchmark uses tiles \#1--\#5 with 10 grasps per tile,
yielding $10 \times 5 = 50$ episodes per approach. Pilot trials on
tiles \#6--\#9 revealed two issues:
\begin{itemize}
  \item \textbf{IK pipeline}: tiles \#6--\#9 (closer to the robot base)
        require the operator to reach downward and behind, changing
        the hand orientation relative to the egocentric camera such
        that the thumb and index finger become partially self-occluded,
        preventing reliable gripper-angle computation.
  \item \textbf{VLA policies}: the robot's own gripper could occlude the
        cube in the policy input view at certain grid positions
        (discussed further in Section~\ref{sec:successrates}).
\end{itemize}
We therefore excluded tiles \#6--\#9 from the final evaluation.

\subsection{VLA Policy Training}
\label{sec:vlatraining}

For the imitation-learning comparison, 50 demonstration episodes
($\times$40\,s each, tiles \#1--\#5) are collected via
leader--follower teleoperation using a second SO-ARM101 as the leader
arm (standard LeRobot teleoperation setup) while the same external
RealSense camera records RGB and depth at 640$\times$480, 30\,FPS.
The demonstration data includes synchronised RGB, depth, and
joint-angle recordings, uploaded to the Hugging Face Hub
in LeRobot dataset format.

Four policies are fine-tuned on Google Colab (T4 GPU, 16\,GB):
\begin{itemize}
  \item \textbf{ACT}~\cite{zhao2023act}: batch size 4--6,
        50\,k steps. Learns a generative model over action chunks
        conditioned on image and proprioceptive observations.
  \item \textbf{SmolVLA}~\cite{cadena2025smolvla}: batch size 64,
        20\,k steps. A 450M-parameter vision-language-action model.
  \item \textbf{$\pi_{0.5}$}~\cite{black2024pi0}: batch size 32,
        3\,k steps, bfloat16. Fine-tuned from a pretrained VLM backbone.
  \item \textbf{GR00T~N1.5}: batch size 32, 3\,k steps. A
        cross-embodiment foundation model.
\end{itemize}

% ============================= 6  RESULTS ===================================
\section{Results and Discussion}
\label{sec:results}

\subsection{Pipeline Latency}

\begin{figure}[t]
\centering
\includegraphics[width=0.85\columnwidth]{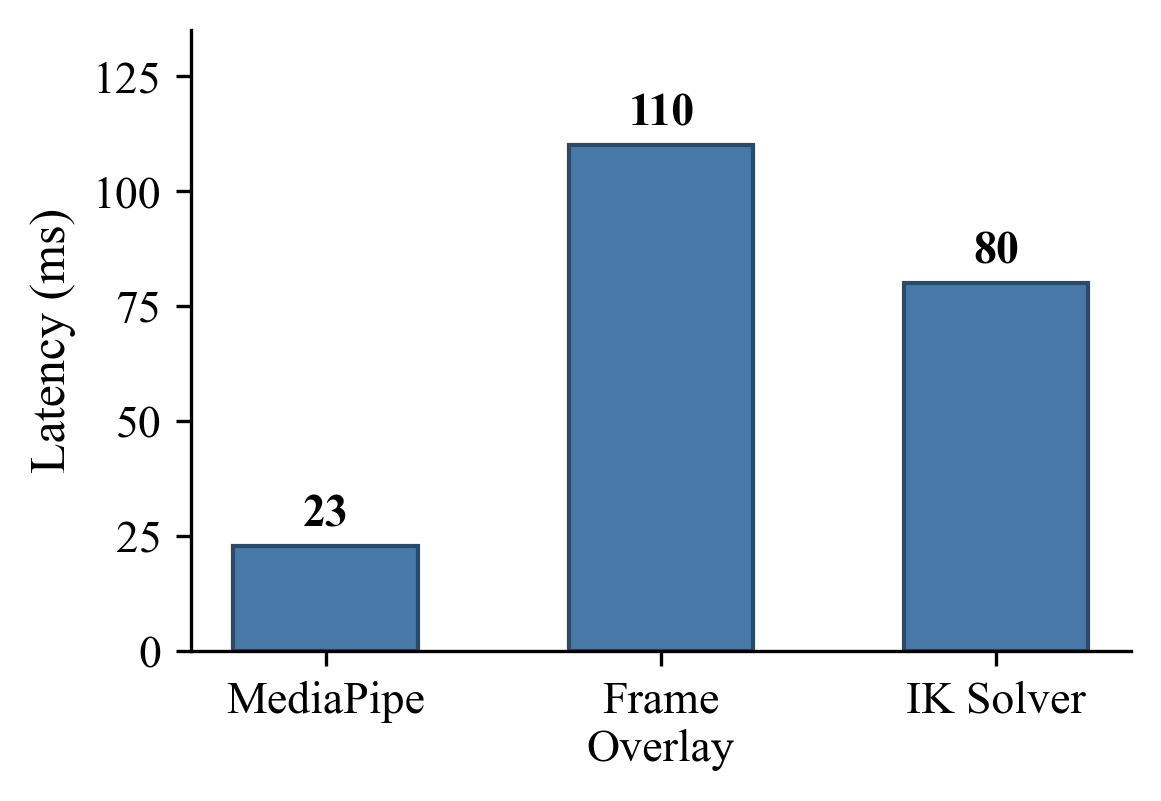}
\caption{\textbf{Per-stage latency breakdown.} The RGB/depth frame overlay
  (110\,ms) and PyBullet IK solver (80\,ms) dominate, yielding a
  total of 213\,ms per frame (${\sim}$5\,FPS). Processing is
  performed offline on recorded \texttt{.bag} files.}
\label{fig:latency}
\end{figure}

Fig.~\ref{fig:latency} shows the per-stage latency breakdown.
The three dominant stages are MediaPipe hand detection (23\,ms),
RGB/depth frame overlay and visualisation (110\,ms), and PyBullet
inverse kinematics solving (80\,ms). The total per-frame processing
time is $23 + 110 + 80 = 213$\,ms, yielding an effective throughput
of approximately ${\sim}$5\,FPS. The pipeline therefore
does \emph{not} operate in real time at the camera's native 30\,FPS:
frames are recorded to a \texttt{.bag} file at 30\,FPS, then processed
offline at ${\sim}$5\,FPS to produce joint-angle trajectories, which
are subsequently replayed on the robot at the target frame rate.

This offline design is deliberate: the pipeline serves as a trajectory
\emph{generation} tool for data collection, not a real-time
controller.

A key advantage of offline processing is that the pipeline can extract
robot joint trajectories from \emph{any} pre-recorded RGB-D video
captured with a known camera pose relative to the robot.
Most imitation-learning frameworks require paired video and action
data, which is typically expensive to collect via leader--follower
teleoperation because it demands two synchronised robot arms.
By contrast, this pipeline generates action labels from human
demonstrations recorded with a simple wearable camera, without
requiring any robot hardware during the recording phase---only
at replay time. Future work could generalise this to a moving
camera by incorporating simultaneous localisation and mapping (SLAM).

\subsection{Pick-and-Place Success Rates}
\label{sec:successrates}

Table~\ref{tab:successrates} reports the pick-and-place success rates
on tiles \#1--\#5. The IK retargeting pipeline was evaluated across
three independent runs of 50 episodes each (150 total episodes),
achieving $\mathbf{86.7\% \pm 4.2\%}$ success with zero training,
since it analytically maps every operator hand motion to the robot.
Its failures stem exclusively from thumb/index landmark detection issues
at certain hand orientations, which prevent the gripper angle from
being computed (the fallback hierarchy recovers in most but not all
cases). In addition to hand occlusion by surrounding objects, finger
visibility relative to the egocentric camera plays a key role: failed
tiles correspond to hand poses where the fingers are aligned with the
camera axis, causing the palm and back of the hand to partially occlude
the thumb and index finger from the camera's perspective.

\begin{figure}[t]
\centering
\includegraphics[width=0.85\columnwidth]{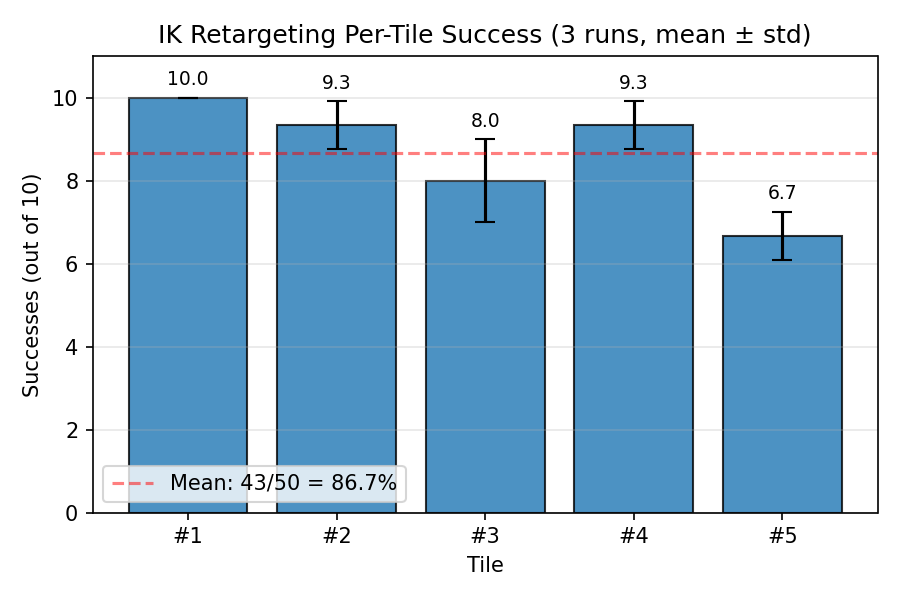}
\caption{\textbf{IK retargeting per-tile success} (out of 10 grasps each,
  averaged across 3 runs).
  Tile~\#1 (farthest from robot, natural hand pose) achieves
  a perfect score. Performance degrades toward tile~\#5 (closest
  to robot base) where the operator's hand orientation causes
  partial thumb/index self-occlusion from the egocentric camera.
  Mean total: 130/150 = 86.7\%.}
\label{fig:pertile}
\end{figure}

Fig.~\ref{fig:pertile} shows the per-tile breakdown for the IK
pipeline, averaged across three runs. Tile~\#1 (farthest from the
robot, most natural hand orientation) achieves a perfect 10/10 across
all runs. Success degrades toward tile~\#5 (closest to the robot base,
requiring downward hand orientation) at $6.7 \pm 0.6$ out of 10,
consistent with the hand self-occlusion failure mode identified in
Section~\ref{sec:task}.

The four VLA policies are trained on 50 leader--follower teleoperation
episodes and evaluated on the same tile set. ACT achieves a
\textbf{92\%} success rate---exceeding the IK pipeline's mean of
86.7\%. One possible explanation is that leader--follower demonstrations
provide physically successful trajectories, whereas the IK pipeline
must reconstruct the motion from vision at every frame and can fail
when hand landmarks are poorly detected. ACT may also benefit from
the larger training budget (50\,k steps) on this structured task.
We note that the IK pipeline requires \emph{zero training} and
serves as a zero-shot baseline; a direct comparison of training
requirements is given in Table~\ref{tab:comparison}.
SmolVLA reaches 50\%, competitive
given its language-conditioning capability but limited by the
small number of training episodes. $\pi_{0.5}$ achieves 40\%, limited
by its short fine-tuning (3\,k steps) from the generalist pretrained
checkpoint. GR00T~N1.5 reaches 35\%, consistent with the challenge of
adapting a cross-embodiment foundation model to a specific low-cost arm.

A likely failure mode for SmolVLA, $\pi_{0.5}$, and GR00T~N1.5
is \emph{self-occlusion}: during the approach phase, the robot's
gripper can partially hide the cube in the camera view, reducing
visibility of the grasp target at a critical moment. ACT appears more
robust to this issue, possibly because action chunking can carry the
policy through short periods of occlusion, although this may also make
the policy more task-specific.

\begin{table}[t]
\caption{\textbf{Pick-and-Place Success Rates} (tiles \#1--\#5, 10 grasps
  per tile). The IK pipeline requires no training and is evaluated
  across 3 runs (150 episodes); VLA policies are trained on
  leader--follower teleoperation demonstrations (50 episodes each).}
\label{tab:successrates}
\centering
\begin{tabular}{lccr}
\hline
\textbf{Approach} & \textbf{Training} & \textbf{Inference} & \textbf{Success} \\
                   & \textbf{Steps}    & \textbf{Hz}        & \textbf{Rate} \\
\hline
IK Retargeting (ours) & ---           & ${\sim}$5 (offline) & $86.7 \pm 4.2$\% \\
ACT~\cite{zhao2023act}          & 50\,k  & ${\sim}$10 & \textbf{92\%} \\
SmolVLA~\cite{cadena2025smolvla} & 20\,k  & ${\sim}$10 & 50\% \\
$\pi_{0.5}$~\cite{black2024pi0} & 3\,k   & ${\sim}$10 & 40\% \\
GR00T~N1.5            & 3\,k          & ${\sim}$10 & 35\% \\
\hline
\end{tabular}
\end{table}

\subsection{Approach Comparison}

Table~\ref{tab:comparison} compares the IK pipeline with the learned
policy approaches across key dimensions.

\begin{table}[t]
\caption{\textbf{Comparison of Teleoperation Approaches}}
\label{tab:comparison}
\centering
\begin{tabular}{p{1.8cm}p{1.4cm}p{3.5cm}}
\hline
\textbf{Property} & \textbf{IK (ours)} & \textbf{Learned Policies} \\
\hline
Training data
  & None
  & 50 episodes (33\,min) \\
Training time
  & 0
  & 2--8\,h on T4 GPU \\
Inference rate
  & ${\sim}$5\,Hz (offline)
  & ${\sim}$10\,Hz \\
Generalisation
  & Task-agnostic
  & Task-specific (ACT) or language-conditioned \\
Failure mode
  & Hand occlusion
  & Gripper self-occlusion \\
\hline
\end{tabular}
\end{table}

\subsection{In-the-Wild Evaluation}
\label{sec:inthewild}

To assess the generality of the IK retargeting approach beyond the
structured lab bench, we deployed the pipeline on the same pick-and-place
task in two unstructured real-world environments: a grocery store and a
pharmacy. The robot and supporting compute setup were placed on a
standard shopping basket, which provides a repeatable height across
trials. Because the SO-ARM101's reach (${\sim}$300\,mm) is roughly half
that of a human arm (${\sim}$600\,mm), the basket had to be positioned
so that target objects on the shelf were within the robot's reachable
volume---typically within ${\sim}$300\,mm of the gripper. This
deliberate placement ensures that failures in the in-the-wild
evaluation reflect pipeline limitations (primarily hand occlusion)
rather than hardware reach constraints.

\begin{figure}[t]
\centering
\includegraphics[width=0.48\columnwidth]{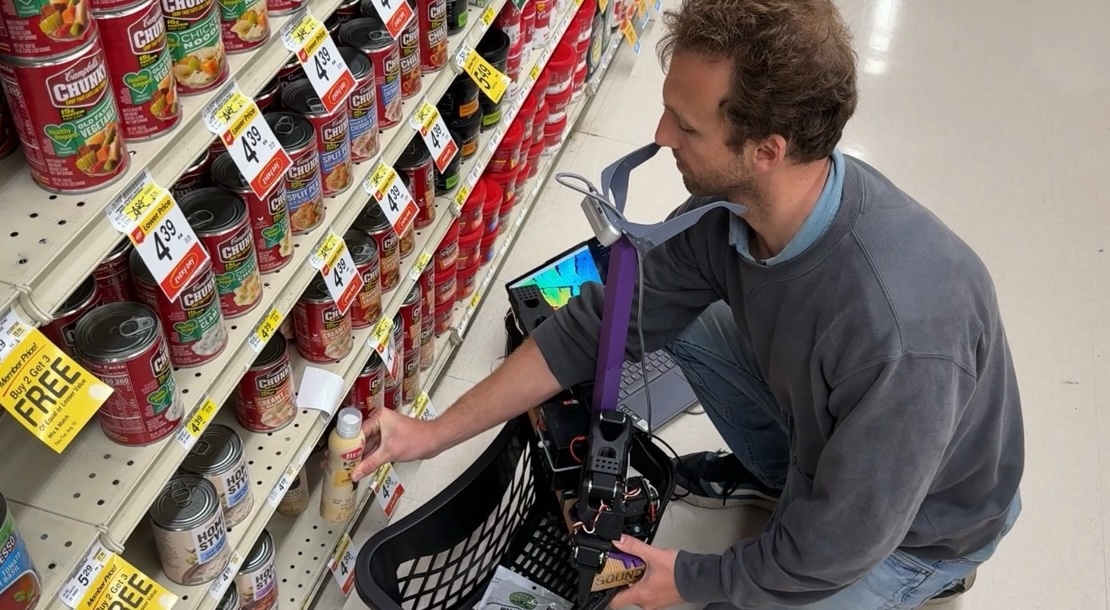}
\hfill
\includegraphics[width=0.48\columnwidth]{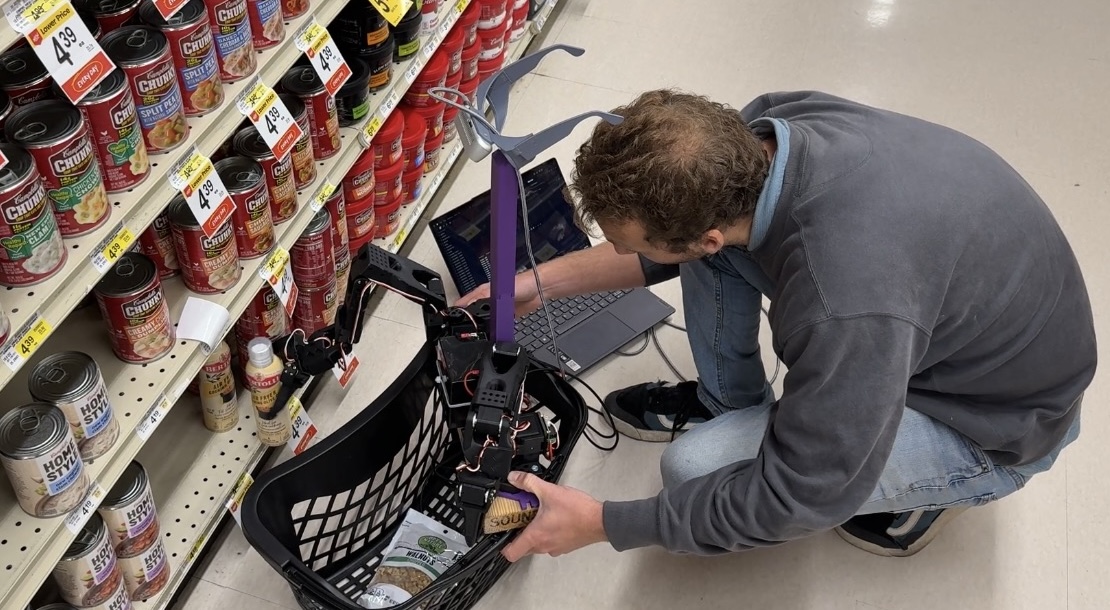}
\caption{\textbf{In-the-wild deployment in a grocery store.} The robot and
  supporting compute setup are placed on a shopping basket for
  repeatable height. Left: the operator reaches for shelf items while
  wearing the glasses-mounted camera.
  Right: the robot mirrors the motion; the PyBullet preview is visible
  on the laptop screen.}
\label{fig:inthewild_setup}
\end{figure}

Fig.~\ref{fig:inthewild_setup} shows the in-the-wild setup: the
operator, still wearing the glasses-mounted camera, reaches for items
on the store shelf (left) while the robot
mounted in the shopping basket reproduces the motion via IK retargeting
after offline processing (right). The task involves grasping various
store items (cans, bottles, boxes) from the shelf and placing them in
the basket.

Over 75 grasp attempts across both locations, only 7 succeeded,
yielding a success rate of \textbf{9.3\%}.

The primary failure mode is \emph{hand occlusion by surrounding
objects}: in a cluttered shelf environment, the operator's hand is
frequently occluded by adjacent products, shelf edges, and price tags
from the egocentric camera's perspective. This causes MediaPipe to lose
track of the thumb and index finger landmarks, preventing both the
gripper angle computation and the IK target position calculation.
Our hand detection reliability analysis
(Section~\ref{sec:handstats}) shows that even in the structured lab
environment, 17.1\% of frames yield no hand detection; in unstructured
scenes with heavy clutter, this rate is expected to be substantially
higher.

\begin{figure}[t]
\centering
\includegraphics[width=0.9\columnwidth]{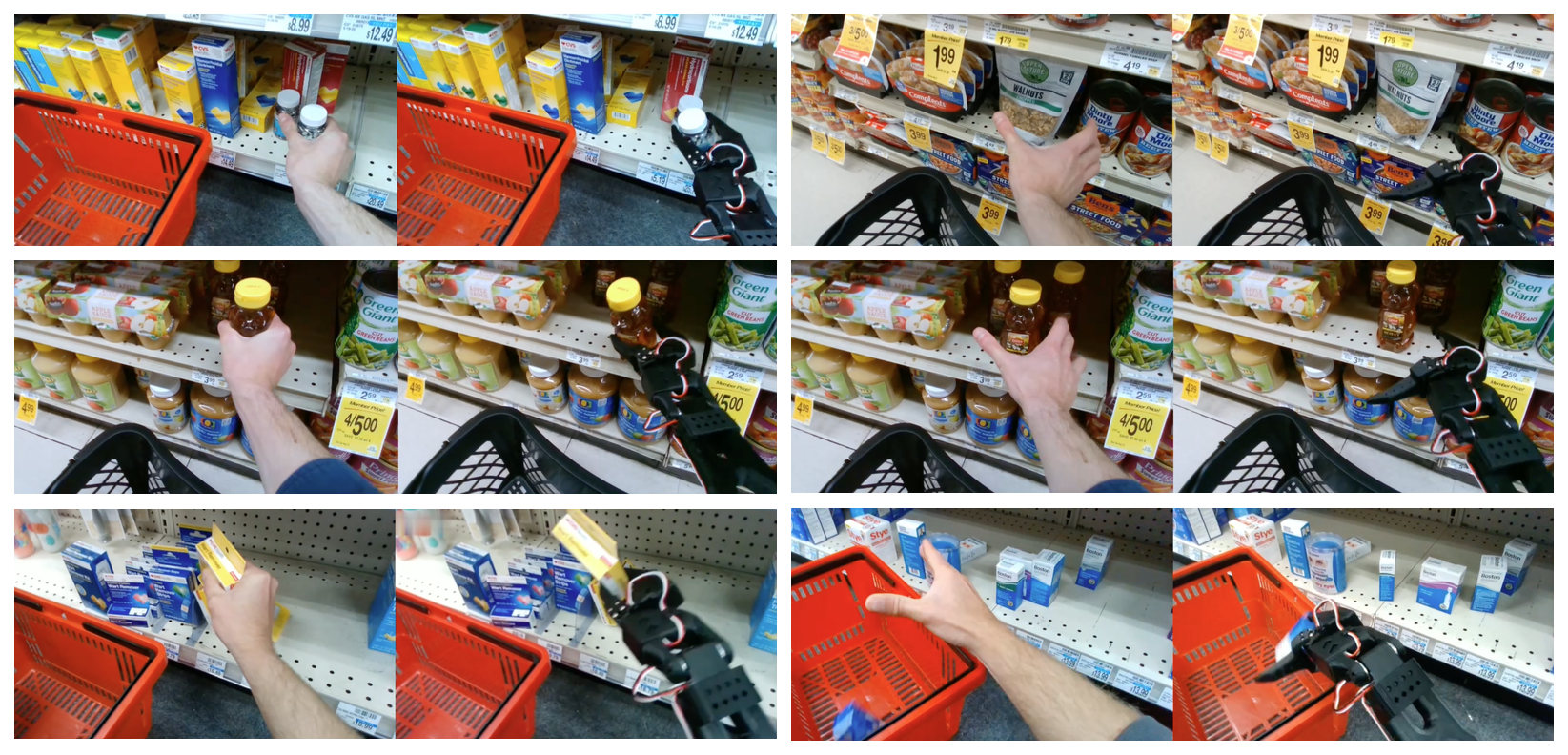}
\caption{\textbf{Mosaic of successful in-the-wild grasps} (7 out of 75
  attempts). Each pair shows the human hand capture (left) and the
  corresponding robot IK retargeting (right) for various store items
  across grocery and pharmacy environments. Despite an overall 9.3\%
  success rate, successful transfers demonstrate the pipeline's
  potential in unstructured scenes.}
\label{fig:inthewild_mosaic}
\end{figure}

Despite the low overall success rate, the 7 successful grasps---shown
in the mosaic in Fig.~\ref{fig:inthewild_mosaic}---indicate that the
pipeline can transfer to unstructured environments when hand visibility
is preserved. As an initial mitigation, we evaluate WiLoR as an
alternative hand detector in Section~\ref{sec:occlusion}, obtaining
an 8\% improvement in detection rate over MediaPipe. Further
mitigation strategies include temporal Kalman filtering to interpolate
through brief occlusion episodes, multi-camera setups to provide
alternative viewpoints, and learned depth completion to handle partial
depth failures.

\subsection{IK Solver Accuracy}
\label{sec:ikaccuracy}

To quantify the accuracy of the IK solver, we re-ran the full pipeline
on a recorded pick-and-place demonstration (274 frames) and compared the
forward kinematics of the solved joint angles $\text{FK}(\vect{q}^*)$
against the original target pose
$(\vect{p}_{\text{target}}, \vect{q}_{\text{target}})$ for each frame
where both were available ($N = 125$).

\begin{figure}[t]
\centering
\includegraphics[width=0.95\columnwidth]{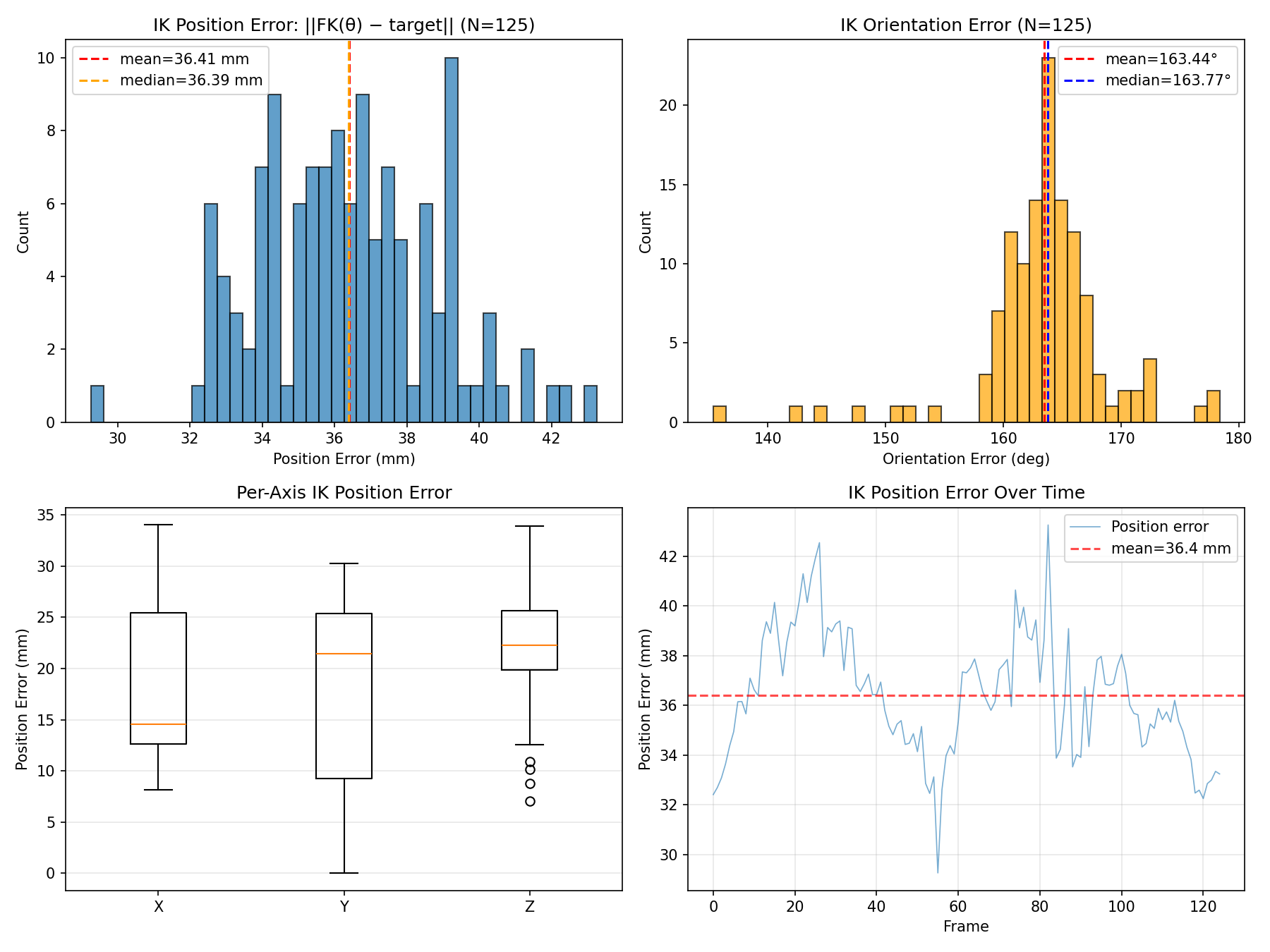}
\caption{\textbf{IK solver accuracy} from real pipeline data ($N=125$
  frames). Top-left: position error histogram.
  Top-right: orientation error histogram.
  Bottom-left: per-axis position error.
  Bottom-right: position error over time.}
\label{fig:ik_error}
\end{figure}

The mean position error $\|\text{FK}(\vect{q}^*) -
\vect{p}_{\text{target}}\|$ is \textbf{36.4\,mm} (std 2.5\,mm,
max 43.3\,mm), distributed approximately equally across axes
(X: 18.4\,mm, Y: 18.2\,mm, Z: 22.7\,mm). The mean orientation error
is $163.4^{\circ}$ (std $5.6^{\circ}$), reflecting the fundamental
constraint of a 5-DOF kinematic chain attempting to reach a 6-DOF
target: because only 5 arm joints contribute to end-effector positioning,
the solver cannot independently satisfy both position and orientation
and must compromise, prioritising position accuracy.
The position error of 36.4\,mm corresponds to approximately 12\% of
the arm's ${\sim}$300\,mm reach, which is acceptable for the grasp
apertures used in our pick-and-place task ($50 \times 50$\,mm cube).
Fig.~\ref{fig:ik_error} shows the error distributions.

\subsection{Trajectory Smoothness}
\label{sec:smoothness}

We evaluate trajectory smoothness using the root-mean-square (RMS) jerk
(third time-derivative of joint angles) computed over 830 valid frames
across 10 contiguous segments from 4 pick-and-place demonstrations
at ${\sim}$5\,FPS.

Table~\ref{tab:smoothness} reports the per-joint RMS jerk for both
the raw IK output and the EMA-smoothed output
($\alpha_{\text{IK}} = 0.5$).
EMA smoothing reduces jerk by 57--68\% across all joints, with the
largest reduction on shoulder pan ($-68\%$) and the smallest on
wrist roll ($-57\%$). The wrist roll and gripper joints exhibit
the highest raw jerk, consistent with the sensitivity of the IK solver
to small landmark perturbations at the end of the kinematic chain.

\begin{table}[t]
\caption{\textbf{Trajectory Smoothness: RMS Jerk by Joint} (rad/s$^3$).
  EMA smoothing ($\alpha_{\text{IK}} = 0.5$) reduces jerk by 57--68\%.}
\label{tab:smoothness}
\centering
\begin{tabular}{lrrr}
\hline
\textbf{Joint} & \textbf{Raw} & \textbf{Smoothed} & \textbf{Reduction} \\
\hline
Shoulder pan  & 10.4 & 3.3 & $-68\%$ \\
Shoulder lift & 16.9 & 6.5 & $-62\%$ \\
Elbow flex    & 25.8 & 9.5 & $-63\%$ \\
Wrist flex    & 35.9 & 13.6 & $-62\%$ \\
Wrist roll    & 47.3 & 20.4 & $-57\%$ \\
Gripper       & 66.2 & 24.2 & $-63\%$ \\
\hline
\end{tabular}
\end{table}

\subsection{EMA Smoothing Ablation}
\label{sec:ablation}

To assess the sensitivity of the pipeline to the IK smoothing
parameter, we varied $\alpha_{\text{IK}}$ over $\{0.3, 0.5, 0.7, 0.8,
0.9, 0.95, 1.0\}$ (where $1.0$ corresponds to no smoothing) while
holding the landmark smoothing parameter fixed at
$\alpha_{\text{lm}} = 0.8$. We measured the resulting trajectory
quality on the same 830-frame dataset.

\begin{figure}[t]
\centering
\includegraphics[width=0.95\columnwidth]{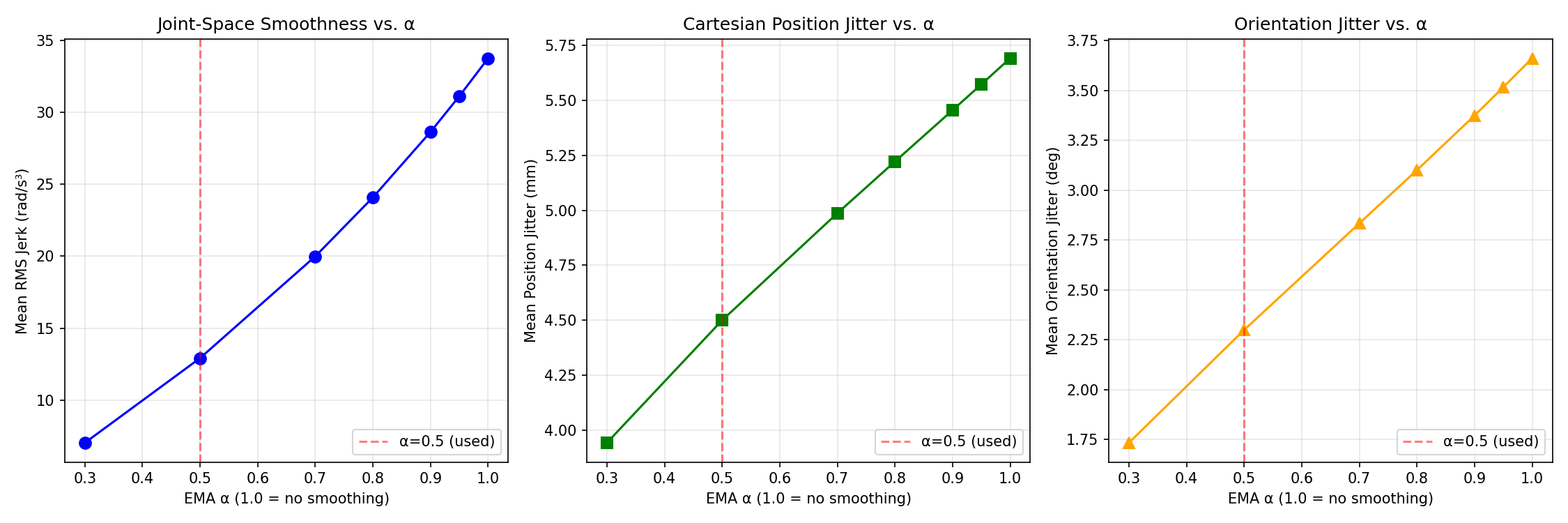}
\caption{\textbf{EMA smoothing ablation.} Joint-space jerk (left),
  Cartesian position jitter (centre), and orientation jitter (right)
  as a function of $\alpha_{\text{IK}}$ (with $\alpha_{\text{lm}} = 0.8$
  fixed). The dashed line marks $\alpha_{\text{IK}} = 0.5$ (used in the
  pipeline).}
\label{fig:ablation}
\end{figure}

Fig.~\ref{fig:ablation} shows the trade-off: as $\alpha_{\text{IK}}$
increases, the pipeline becomes more responsive to hand motion but at
the cost of increased jerk and Cartesian jitter. At
$\alpha_{\text{IK}} = 0.3$, mean RMS jerk is 7.0\,rad/s$^3$ and
position jitter is 3.9\,mm, but the trajectory lags behind the
operator's hand. At $\alpha_{\text{IK}} = 1.0$ (no smoothing), jerk
rises to 33.7\,rad/s$^3$ and jitter to 5.7\,mm.
The chosen value of $\alpha_{\text{IK}} = 0.5$ provides a balanced
trade-off: 12.9\,rad/s$^3$ jerk, 4.5\,mm position jitter, and
2.3$^{\circ}$ orientation jitter.

\subsection{End-Effector Trajectory Analysis}
\label{sec:eetrajectory}

We computed the frame-to-frame end-effector position and orientation
changes (Cartesian jitter) via forward kinematics over 839 frames from
4 pick-and-place trajectories.
The mean frame-to-frame position change is 5.75\,mm
(smoothed: 4.52\,mm, $-21\%$), and the mean orientation change is
$3.89^{\circ}$ (smoothed: $2.42^{\circ}$, $-38\%$).
The end-effector workspace spans
$306 \times 449 \times 299$\,mm, consistent with the SO-ARM101's
${\sim}$300\,mm reach.

\begin{figure}[t]
\centering
\includegraphics[width=0.95\columnwidth]{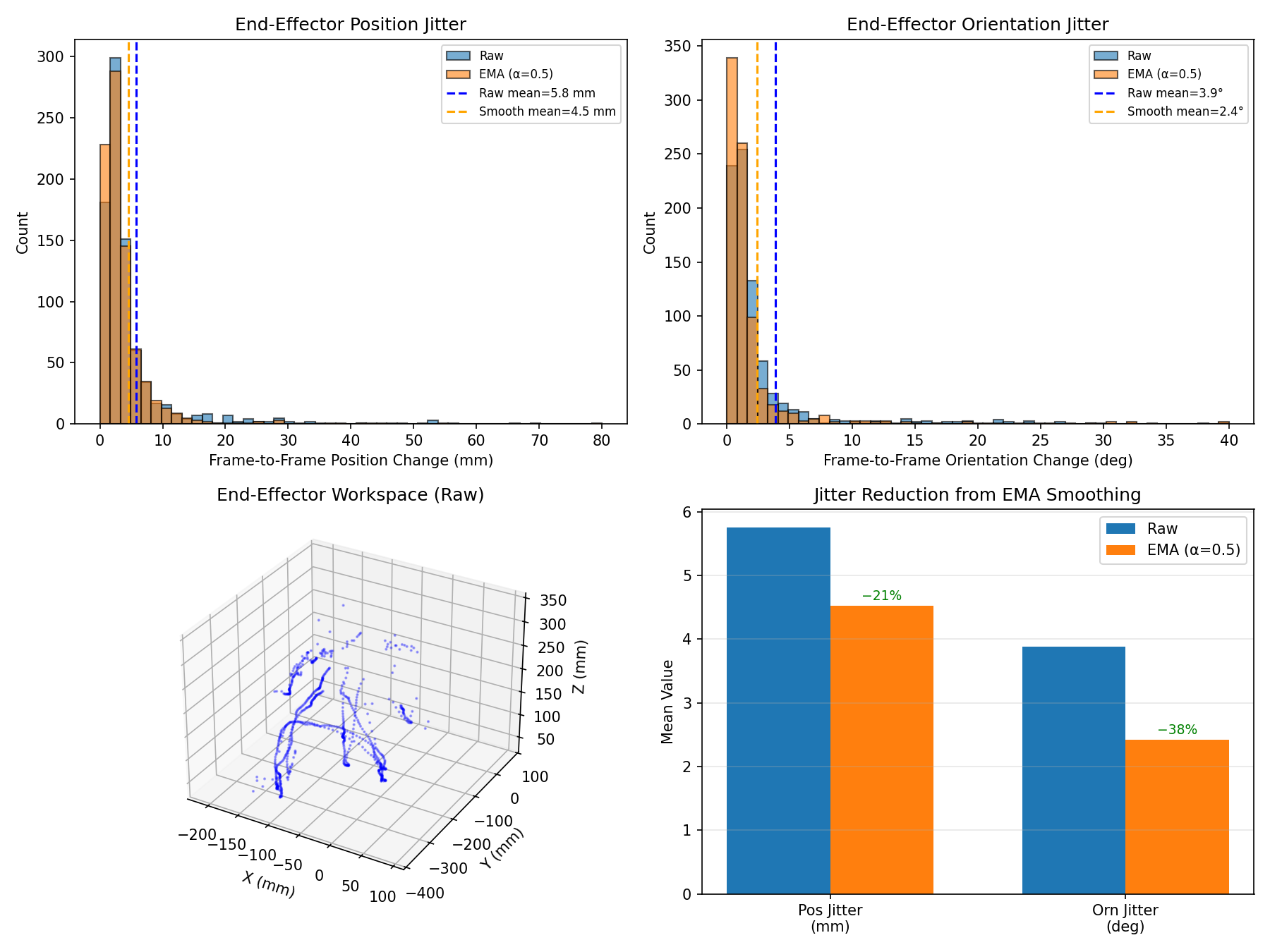}
\caption{\textbf{End-effector trajectory analysis.}
  Top: position and orientation jitter distributions (raw vs.\
  EMA-smoothed). Bottom-left: 3D workspace scatter.
  Bottom-right: jitter reduction summary.}
\label{fig:ee_accuracy}
\end{figure}

\subsection{Joint Saturation Analysis}
\label{sec:saturation}

Table~\ref{tab:saturation} reports the percentage of frames where each
joint operates near its URDF-defined limits (within 0.05\,rad).
Overall saturation is low ($<5\%$), indicating that the IK solver
rarely encounters workspace boundaries during typical pick-and-place
trajectories. The wrist flex joint shows the highest saturation
($4.7\%$ at upper limit), followed by wrist roll ($2.1\%$) and
elbow flex ($1.9\%$ at lower limit).

\begin{table}[t]
\caption{\textbf{Joint Saturation} (\% of 855 frames within 0.05\,rad
  of URDF joint limits).}
\label{tab:saturation}
\centering
\begin{tabular}{lrrr}
\hline
\textbf{Joint} & \textbf{At Lower} & \textbf{At Upper} & \textbf{Total} \\
\hline
Shoulder pan  & 0.1\% & 0.0\% & 0.1\% \\
Shoulder lift & 0.2\% & 0.0\% & 0.2\% \\
Elbow flex    & 1.9\% & 0.0\% & 1.9\% \\
Wrist flex    & 0.0\% & 4.7\% & 4.7\% \\
Wrist roll    & 0.0\% & 2.1\% & 2.1\% \\
Gripper       & 0.0\% & 0.0\% & 0.0\% \\
\hline
\end{tabular}
\end{table}

\subsection{Hand Detection Reliability}
\label{sec:handstats}

Over 503 frames of a representative pick-and-place recording,
MediaPipe detected the right hand in 77.3\% of frames, with 5.6\%
erroneously classified as left-hand detections and 17.1\% yielding
no detection. After applying an EMA-based hand-swap correction
heuristic, erroneous left-hand classifications were reduced from
5.6\% to 2.8\%, while the no-detection rate remained at 17.1\%
(since no heuristic can recover frames where no hand is visible).
The 17.1\% no-detection rate is consistent with the structured-environment
failure modes: frames where the operator's hand orientation causes
partial self-occlusion from the egocentric camera. Replacing MediaPipe
with WiLoR reduces the no-detection rate from 37.2\% to 32.3\% on
a separate 347-frame recording (Section~\ref{sec:occlusion}),
confirming that the hand localiser is the detection bottleneck.

\subsection{Sim-to-Real Transfer}
\label{sec:simreal}

\begin{figure}[t]
\centering
\includegraphics[width=0.85\columnwidth]{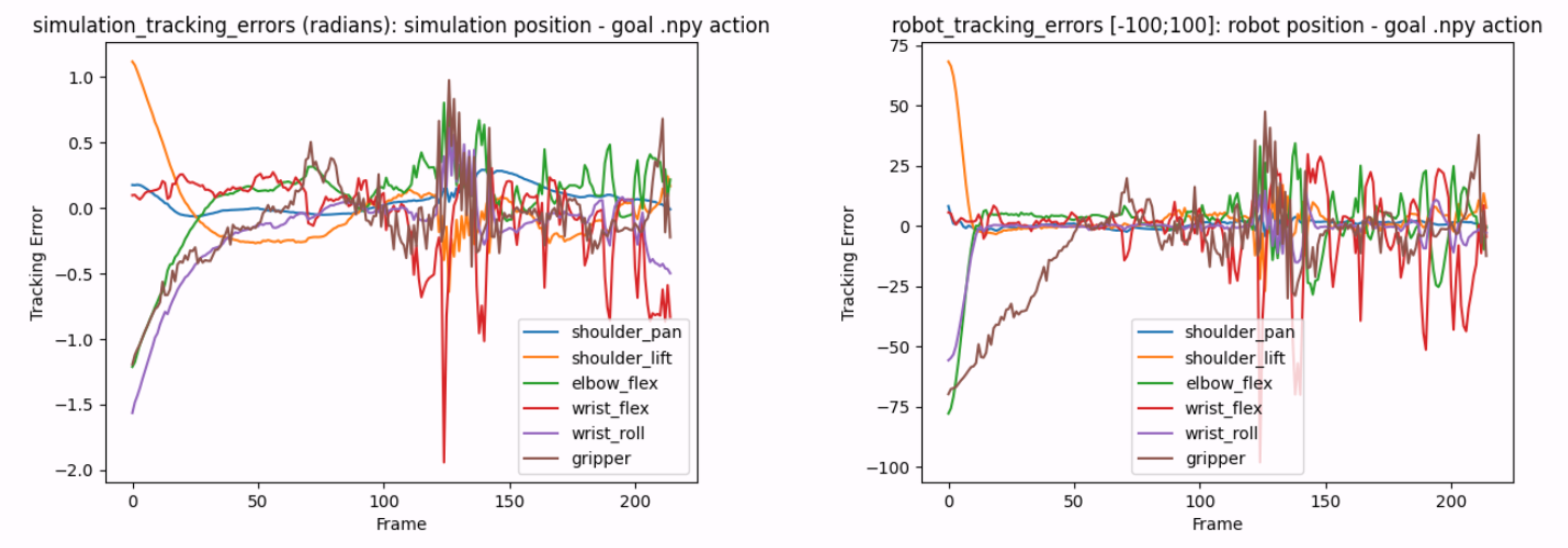}
\caption{\textbf{Sim-to-real tracking error.} Per-joint tracking error
  (commanded minus achieved position, in motor-space units $[-100, 100]$)
  during physical robot replay. The error oscillates around zero,
  confirming unbiased tracking. The gripper exhibits the
  largest transient errors.}
\label{fig:simreal}
\end{figure}

We validated the fidelity of the PyBullet simulation by comparing
the commanded joint trajectories against the physical robot's
achieved positions during replay. Fig.~\ref{fig:simreal} shows the
per-joint tracking error (commanded position minus achieved position)
over a representative pick-and-place trajectory.
The tracking error oscillates around zero (unbiased), confirming
that the PID tuning (P$=$12, I$=$0, D$=$24) accurately reproduces
the commanded trajectories on the physical hardware.
The high-frequency noise visible in the error signal originates from
the commanded actions themselves, which inherit jitter from the
MediaPipe landmark detection and IK solving stages. Because the
oscillation period is on the order of a few frames and the error
remains centred around zero, the physical robot is correctly smoothing
through this jitter rather than tracking it---a desirable property of
the servo PID configuration. The gripper joint exhibits the largest transient errors, attributable
to the higher jerk in commanded trajectories for this joint
(consistent with the smoothness analysis in
Section~\ref{sec:smoothness}) and the inherent servo lag of the
gripper actuator.

\subsection{Occlusion Mitigation via WiLoR}
\label{sec:occlusion}

The dominant failure mode is MediaPipe's inability to detect the hand
when it is occluded by surrounding objects or when the hand pose causes
the fingers to be occluded by the palm from the camera's perspective
(Section~\ref{sec:inthewild}). MediaPipe Hands uses a two-stage
architecture: the BlazePalm detector first localises a palm bounding
box, then a landmark regressor predicts 21 keypoints within that crop.
When the palm is partially occluded, the BlazePalm detector fails
entirely and no landmarks are produced---the pipeline receives zero
information for that frame.

To investigate whether a more robust hand detector can recover these
lost frames, we integrated WiLoR~\cite{potamias2024wilor}, which
replaces BlazePalm with a DarkNet-based YOLO localiser followed by a
ViT-based 3D hand reconstructor trained on the MANO hand
model~\cite{romero2017mano}. We ran both detectors on a 347-frame
egocentric pick-and-place recording from the structured lab environment
and compared detection rate (percentage of frames where the hand is
detected) and valid IK target rate (percentage of frames where a
usable position and orientation can be computed from the detected
landmarks plus depth).

Table~\ref{tab:occlusion} summarises the results. WiLoR detects the
hand in 67.7\% of frames compared to MediaPipe's 62.8\%, an improvement
of 7.8\%. The valid IK target rate follows a similar trend: 64.8\% for
WiLoR versus 59.9\% for MediaPipe (+8.2\%). In 3.5\% of frames, WiLoR
successfully detected the hand where MediaPipe produced no detection at
all, demonstrating partial recovery of occluded frames. Notably, both
detectors convert detected hands to valid IK targets at high rates
(95.7\% for WiLoR, 95.4\% for MediaPipe), confirming that the IK target
computation itself is not the bottleneck---hand detection is.

\begin{table}[t]
\centering
\caption{Hand detector comparison on 347 egocentric frames (left hand,
structured lab environment).}
\label{tab:occlusion}
\begin{tabular}{lcc}
\toprule
\textbf{Metric} & \textbf{MediaPipe} & \textbf{WiLoR} \\
\midrule
Hand detected        & 218 (62.8\%) & 235 (67.7\%) \\
Valid IK target      & 208 (59.9\%) & 225 (64.8\%) \\
No detection         & 129 (37.2\%) & 112 (32.3\%) \\
\midrule
Recovery (WiLoR found, MP missed) & \multicolumn{2}{c}{12 frames (3.5\%)} \\
Detection improvement & \multicolumn{2}{c}{+7.8\%} \\
Valid target improvement & \multicolumn{2}{c}{+8.2\%} \\
\bottomrule
\end{tabular}
\end{table}

The improvement is modest in the structured lab environment, where
occlusion is primarily caused by self-occlusion of the operator's hand
rather than external objects. In unstructured environments with heavy
clutter (shelves, products, price tags), the gap is expected to widen
because DarkNet's object-detection training makes it more robust to
partial occlusion than BlazePalm's palm-specific detector.

The trade-off is computational: MediaPipe is lightweight and CPU-only,
whereas WiLoR's DarkNet + ViT pipeline is two orders of magnitude
larger and requires GPU inference---acceptable for our offline
${\sim}$5\,FPS budget but the dominant cost for real-time use.
WiLoR is integrated in our repository as a drop-in replacement
for the MediaPipe detector (note: WiLoR is released under a
non-commercial license).

\subsection{Failure Modes}

We identify the following primary failure modes, separated by context.

\subsubsection{Structured Lab Environment}
\begin{enumerate}
  \item \textbf{Thumb/index detection failure}: At certain hand
        orientations (particularly toward tiles
        \#6--\#9), the thumb and index finger landmarks are poorly
        detected, preventing gripper angle computation. This is the
        dominant IK failure mode (accounting for all lab
        failures).
  \item \textbf{Gripper self-occlusion (VLAs)}: The robot's gripper
        can hide the cube in the camera view during the approach
        phase. This effect appears most pronounced for SmolVLA,
        $\pi_{0.5}$, and GR00T~N1.5, where the vision model loses
        sight of the grasp target at a critical moment.
  \item \textbf{Gripper jitter}: Rapid open-close oscillations occur
        when the thumb-index angle is near the binary threshold. The
        binary gripper mode with a $60^{\circ}$ threshold resolves this.
\end{enumerate}

\subsubsection{Unstructured In-the-Wild Environments}
\begin{enumerate}
  \item \textbf{Hand occlusion by scene objects}: Shelf products,
        price tags, and adjacent items frequently occlude the
        operator's hand from the egocentric camera, causing MediaPipe
        to lose landmark tracking. This is the dominant failure mode,
        accounting for the drop from 86.7\% to 9.3\% success.
  \item \textbf{Workspace mismatch}: The human arm (${\sim}$60\,cm)
        reaches further than the robot (${\sim}$30\,cm), requiring
        careful positioning of the shopping basket to keep objects
        within the robot's reachable volume.
  \item \textbf{Depth noise}: Reflective packaging (foil, plastic
        wrap) produces unreliable depth readings, causing spurious
        3D deprojection results.
\end{enumerate}

In practice, an attentive operator can mitigate the first two failure
modes by tilting the wrist to keep thumb and index fingers visible to
the egocentric camera and by pre-positioning the basket so that
candidate items lie within the robot's $\sim$30\,cm reachable volume.

% ============================= 7  CONCLUSION ================================
\section{Conclusion}

We have presented an offline hand-shadowing retargeting pipeline for
robot trajectory generation via egocentric hand tracking and analytical
inverse kinematics. The pipeline processes recorded RGB-D frames at
${\sim}$5\,FPS (213\,ms per frame) using MediaPipe Hands for landmark
detection, depth-based 3D deprojection, and PyBullet for IK solving
and trajectory preview, with the LeRobot framework for physical
deployment on the SO-ARM101.

On a structured pick-and-place benchmark (tiles \#1--\#5,
150 episodes across 3 runs), the IK retargeting pipeline achieves
$86.7\% \pm 4.2\%$ success with zero training. The IK solver achieves
a mean position error of 36.4\,mm, with orientation error dominated by the
5-DOF kinematic constraint. EMA smoothing reduces trajectory jerk by
57--68\%, and an ablation study confirms that $\alpha_{\text{IK}} = 0.5$ provides
a balanced smoothness--responsiveness trade-off.
Among the teleoperation-trained VLA policies, ACT reaches 92\%, while
SmolVLA (50\%), $\pi_{0.5}$ (40\%), and GR00T~N1.5 (35\%) are
substantially lower. In-the-wild evaluation in grocery store and
pharmacy environments (7/75 successful grasps, 9.3\%) indicates that
hand occlusion by surrounding objects is the primary limitation.

\subsection{Future Work}
\begin{itemize}
  \item \textbf{Occlusion-robust hand tracking}: Our WiLoR integration
        (Section~\ref{sec:occlusion}) demonstrates an 8\% detection-rate
        improvement over MediaPipe in structured environments; further
        gains may come from temporal Kalman filtering to interpolate
        through brief occlusion episodes, learned depth completion,
        or multi-camera setups to maintain landmark visibility in
        heavily cluttered environments.
  \item \textbf{6-DoF grasp pose estimation}: Replacing the
        geometric thumb--index angle with a learned grasp-pose
        predictor to handle diverse object shapes and materials.
  \item \textbf{Dual-arm extension}: Extending to simultaneous
        bimanual control.
  \item \textbf{Industrial data collection}: Because the only
        wearable hardware is a pair of camera glasses, the pipeline
        is well suited to capturing demonstrations directly on
        factory or warehouse floors, where leader--follower setups
        are impractical.
\end{itemize}

\section*{Acknowledgment}
The authors thank the UC Berkeley Department of Mechanical Engineering
and the Fung Institute for Engineering Leadership for providing
access to lab facilities, equipment, and student collaboration.

\begin{IEEEbiography}[{\includegraphics[width=1in,height=1.25in,clip,keepaspectratio]{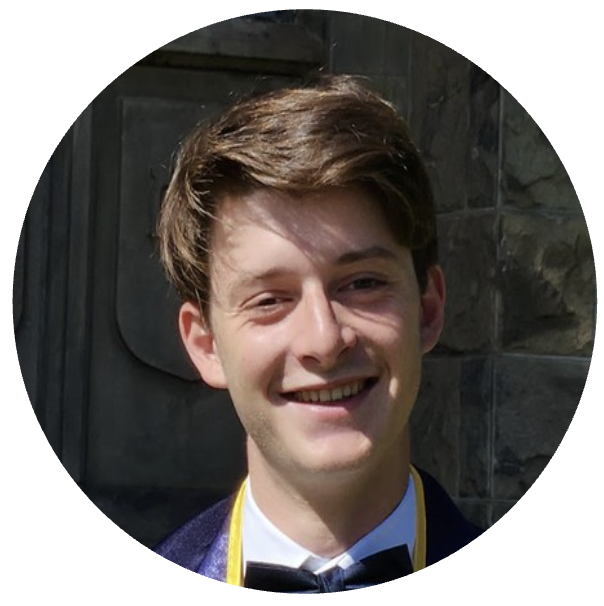}}]{Hendrik Chiche}
is with OMGrab Inc.\ and affiliated with the University of California,
Berkeley through the Capstone Project program.
His research interests include computer vision, robotics, and
human-to-robot skill transfer.
\end{IEEEbiography}

\begin{IEEEbiography}[{\includegraphics[width=1in,height=1.25in,clip,keepaspectratio]{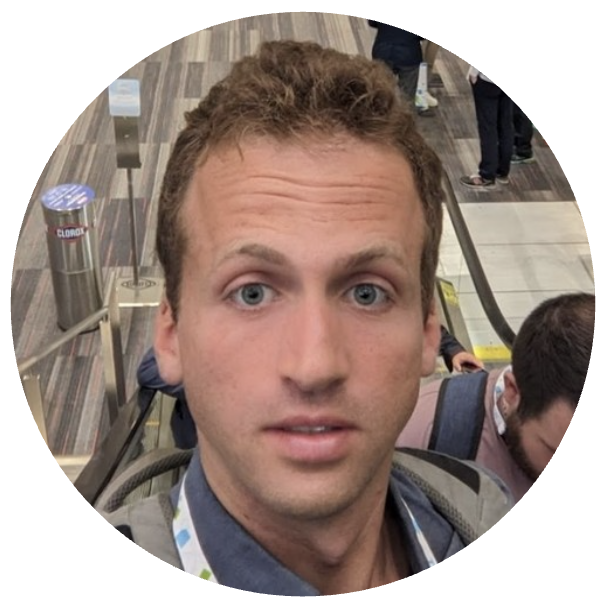}}]{Antoine Jamme}
is with OMGrab Inc.\ and affiliated with the University of California,
Berkeley through the Capstone Project program.
His research interests include robotic manipulation, imitation
learning, and low-cost robotics platforms.
\end{IEEEbiography}

\begin{IEEEbiography}[{\includegraphics[width=1in,height=1.25in,clip,keepaspectratio]{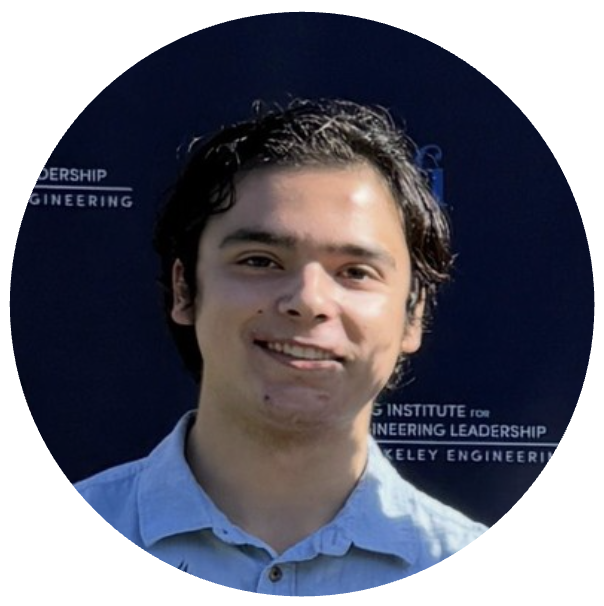}}]{Trevor Rigoberto Martinez}
is a graduate student in the Department of Mechanical Engineering
at the University of California, Berkeley.
\end{IEEEbiography}

\begin{IEEEbiography}[{\includegraphics[width=1in,height=1.25in,clip,keepaspectratio]{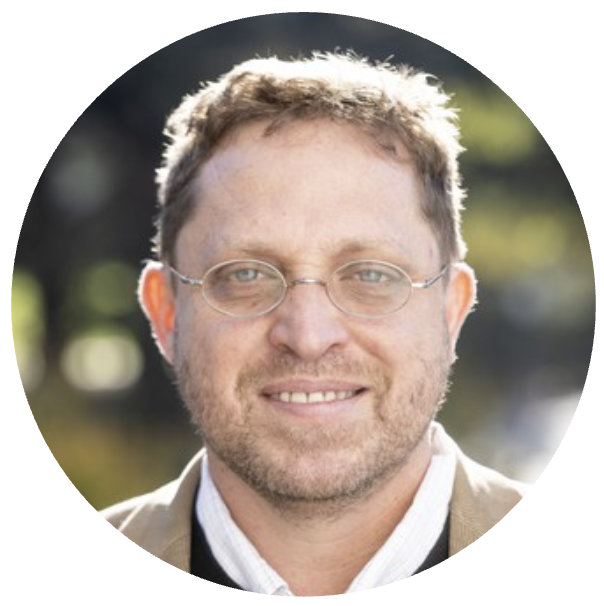}}]{Gabriel Gomes}
is a Lecturer in the Department of Mechanical Engineering at the
University of California, Berkeley. He received a doctorate degree in
automatic control theory from UC Berkeley in 2004. His research
interests include modeling, simulation, and control of engineering
systems.
\end{IEEEbiography}

\end{document}